\documentclass[a4paper, 10pt, journal]{IEEEtran}
\usepackage{amsmath,amssymb,amsfonts}
\usepackage{array}
\usepackage[caption=false,font=normalsize,labelfont=sf,textfont=sf]{subfig}
\usepackage{textcomp}
\usepackage{stfloats}
\usepackage{url}
\usepackage{verbatim}
\usepackage{graphicx}
\usepackage{cite}
\usepackage{multirow}
\usepackage{multicol}
\usepackage{svg}
\usepackage{algpseudocode}
\usepackage{algorithm}
\usepackage{hyperref}
\hypersetup{
    colorlinks=true,
    urlcolor=blue,
}
\usepackage{orcidlink}
\usepackage{float}
\DeclareMathOperator*{\argmin}{argmin}

\newcommand{\TODO}[1]{#1}

\newcommand{\kike}[1]{#1}

\newcommand{\ycheng}[1]{{#1}}
\newcommand{\justin}[1]{{#1}}

\newcommand{\dennis}[1]{{#1}}

\newcommand{\R}[1]{\mathbb{R}}
\newcommand{\note}[1]{}

\begin{document}

\title{360-DFPE: Leveraging Monocular 360-Layouts for Direct Floor Plan Estimation}

% \author{IEEE Publication Technology,~\IEEEmembership{Staff,~IEEE,}
%         % <-this % stops a space
% \thanks{This paper was produced by the IEEE Publication Technology Group. They are in Piscataway, NJ.}% <-this % stops a space
% \thanks{Manuscript received April 19, 2021; revised August 16, 2021.}
% }

\author{
Bolivar Solarte$^{*1}$ \orcidlink{0000-0003-3518-755X}, 
Yueh-Cheng Liu$^{*1}$ \orcidlink{0000-0002-8053-9801}, 
Chin-Hsuan Wu$^{1}$ \orcidlink{0000-0003-1547-0825},
Yi-Hsuan Tsai$^{2}$ \orcidlink{0000-0002-6191-0134},
% \and
% Wei-Chen Chiu$^{2}$\ % ,\orcidlink{0000-0001-7715-8306}
and
Min Sun$^{1}$ \orcidlink{0000-0001-9598-8178}
\thanks{$^1$ National Tsing Hua University}
% % \thanks{$^2$ National Yang Ming Chiao Tung University}
% \thanks{$^2$ Yi-Hsuan Tsai is with Phiar Technologies.}
\thanks{$^2$ Phiar Technologies}
% \thanks{$^3$ NEC Labs America}
\thanks{$^*$ The authors contribute equally to this paper.}
% \thanks{Digital Object Identifier (DOI): see top of this page.}
}

% The paper headers
% \markboth{IEEE ROBOTICS AND AUTOMATION LETTERS. PREPRINT VERSION. ACCEPTED May, 2022}%
% {Solarte \MakeLowercase{\textit{et al.}}: 360-DFPE: Leveraging Monocular 360-Layouts for Direct Floor Plan Estimation}

% \IEEEpubid{0000--0000/00\$00.00~\copyright~2021 IEEE}
% Remember, if you use this you must call \IEEEpubidadjcol in the second
% column for its text to clear the IEEEpubid mark.

\maketitle

\begin{abstract}
We present 360-DFPE, a sequential floor plan estimation method that directly takes 360-images as input without relying on active sensors or 3D information. Our approach leverages a loosely coupled integration between a monocular visual SLAM solution and a monocular 360-room layout approach, which estimate camera poses and layout geometries, respectively. Since our task is to sequentially capture the floor plan using monocular images, the entire scene structure, room instances, and room shapes are unknown. To tackle these challenges, we first handle the scale difference between visual odometry and layout geometry via formulating an entropy minimization process, which enables us to directly align 360-layouts without knowing the entire scene in advance. Second, to sequentially identify individual rooms, we propose a novel room identification algorithm that tracks every room along the camera exploration using geometry information. Lastly, to estimate the final shape of the room, we propose a shortest path algorithm with an iterative coarse-to-fine strategy, which improves prior formulations with higher accuracy and faster run-time. Moreover, we collect a new floor plan dataset with challenging large-scale scenes, providing both point clouds and sequential 360-image information. Experimental results show that our monocular solution achieves favorable performance against the current state-of-the-art algorithms that rely on active sensors and require the entire scene reconstruction data in advance. The code and dataset are available at the project page: \url{https://enriquesolarte.github.io/360-dfpe/}.
\end{abstract}


\section{Introduction}
% the main idea of direct was to use directly the image info w/o active sensor
% % %
%  
%
%
% % %
\kike{
      \IEEEPARstart{T}{he} underlying 3D structures of an indoor scene, such as a floor plan of a building, play a crucial role for a holistic robot perception as detailed in~\cite{davison2018futuremapping, boniardi2017robustLoc, Rosinol20rss-dynamicSceneGraphs, boniardi2019robotLoc, Ravichandran21-RLwithSceneGraphs, hampali2021monte}.
      Despite their simplicity, these high-level geometry abstractions can complement challenging tasks such as obstacle avoidance~\cite{Rosinol20rss-dynamicSceneGraphs}, robot localization~\cite{boniardi2019robotLoc}, path planning~\cite{Ravichandran21-RLwithSceneGraphs}, and scene understanding~\cite{hampali2021monte}; hence, a handy and direct estimation of a floor plan geometry is our primary motivation.
    %   Defining such a high-level geometry abstraction can complement tasks such as obstacle avoidance, robot localization~\cite{boniardi2019robotLoc}, path planning, and scene understanding~\cite{Rosinol20rss-dynamicSceneGraphs}; hence, an handy and direct estimation of a floor plan geometry is our main motivation.
}
\TODO{Current state-of-the-art solutions for floor plan estimation~\cite{chen2019floorSP, liu2018floornet, stekovic2021montefloor} rely on dense point clouds as input, which require active sensors (e.g., LiDAR, depth cameras) for data collection and pre-processing steps beforehand};
% Current state-of-the-art solutions for floor plan estimation~\cite{chen2019floorSP, liu2018floornet, stekovic2021montefloor} mainly rely on active sensors (e.g., LiDAR, depth cameras) to acquire and register multiple 3D measurements as their input data collected beforehand (e.g., point cloud);
% to build their input data in beforehand (e.g., point cloud);
% , requiring a pre-processing stage that registers all 3D measurements into a single point cloud before estimating the floor plan geometry
hence, a direct estimation from a sequence of observation is \TODO{not supported}.
% , e.g., using depth cameras[x][x] and light range sensors (LiDAR)[X].
On the other hand, approaches that rely only on imagery \cite{cabral2014piecewise, pintore20183d} generally leverage structure-from-motion (SfM), multi-view stereo (MVS), and pixel semantic estimations to project geometric clues into 3D/2D space from where the floor plan is inferred. 
% However, this projection requires the whole scene data and knowing the camera height in advance, narrowing their applicability to controlled scenarios and a non-sequential estimation.
\kike{However, these solutions heavily depend on the estimated point cloud's quality and sparsity. Additionally, they require the whole data in advance to successfully assess multiple rooms in the scene.}

% This solution has the capability the process data sequentially, however these solutions mainly use features which additinal projects undesaried feature coming from furniture and texture detals. Hence, it requires a postprocessing to remove, clean, and finally build a floor plan geometry. 
% 
% A branch of Visual Simultaneous Localization and Mapping (VSLAM) systems \cite{yang2019monocularFPE, le2017denseFPE, pumarola2017pl-slam} use point, line, plane, or depth features to \ycheng{sequentially} estimate and register reliable 3D structures \ycheng{from images}. However, these methods usually estimate the 3D scene geometry as several piece-wise planes \ycheng{instead of room walls}, which is likely to contain undesired furniture and texture details.

\kike{Other solutions~\cite{yang2019monocularFPE, le2017denseFPE, pumarola2017pl-slam} leverage Visual Simultaneous Localization and Mapping (VSLAM) by estimating the 3D scene structure using point, line, plane, and depth features from a sequence of images.}
% Another line of work is the \st{A branch of} Visual Simultaneous Localization and Mapping (VSLAM) system \cite{yang2019monocularFPE, le2017denseFPE, pumarola2017pl-slam} that sequentially estimate the 3D scene structure using point, line, plane and depth features from images.
% to \ycheng{sequentially} estimate and register reliable  \ycheng{from images}. However, these methods usually estimate the 3D scene geometry as several piece-wise planes \ycheng{instead of room walls}, which is likely to contain undesired furniture and texture details.
% However, these methods usually represent the  3D scene as several piece-wise plane estimates, which do not represent directly a floor plan structure \TODO{which cannot be served as floor plans.}
\kike{
However, this scene structure is usually represented by a set of piece-wise planes that include all furniture and texture details, without clearly defining room layouts or multiple room instances; hence, it cannot directly represent a floor plan structure.}
% However, the output cannot be served as floor plans due to the lack of room instances.
To the best of our knowledge, an approach that directly uses images to estimate the floor plan geometry sequentially has not been addressed yet.
% Although these solutions require only image information, they still need a pre-processing stage

\begin{figure}

\centering

%%% USE SVG for RA-L submission
% \includesvg[width=0.9\linewidth]{FIG/teaser_v8.svg}
% \[width=0.9\linewidth]{FIG/teaser_v8.png}
% Use PNG for arxiv submission
\includegraphics[width=0.95\linewidth]{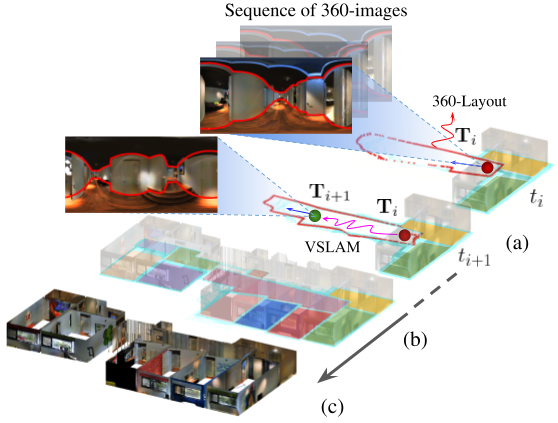}
\vspace{-2mm}
\caption{
\textbf{360-DFPE: Direct Floor Plan Estimation using sequence of 360-images.}
Our goal is to sequentially estimate the floor plan by aggregating the 360 layout estimations (red-line) directly from the images.
(a) We illustrate the process of 360-DFPE under two camera positions ($\mathbf{T}_i$, $\mathbf{T}_{i+1}$) at consecutive times ($t_i$, $t_{i+1}$).
(b) A complete estimation of the floor plan is presented as multiple room shapes (denoted by different colors).
(c) A visually pleasing 3D scene aligned with the estimated floor plan is rendered using the 360-images as texture.
% Note that all the processes are completed in a sequential fashion, leveraging off-the-shelf VSLAM and 360-layout estimator.
Note that only a passive 360-camera is required and the process can be completed in a sequential fashion.
%Note that although this representation is defined by a few planes and points (walls and corners per room), it contents much of the 3D information of the scene.
}\label{fig:teaser}
\vspace{-4mm}
\end{figure}

%%%%%
% Structure.
% 1. Layout prediction single room (normal vs 360)
% 2. Single room layout not accurate because noise, large room...
% 3. Cannot capture the scene relation between rooms
% We can solve these by aggregating sequence of layout
% However, multiple layouts -> challenges
%%%%%

\dennis{
One way to \ycheng{utilize sequential monocular images for floor plan estimation is to} directly apply room layout estimation methods \cite{2020_TransMult_Room_lY, hsiao2019flat2layout_room_ly, Lee_2017_ICCV_room_ly,wang2021led2, Pintore_atlanta_net2020, sun2019horizonnet, sun2021hohonet}. However, even by leveraging large field-of-view images, i.e., 360-images, some issues remain unsolved. For example, the room layout estimation may differ with different camera positions, or it may fail to capture large rooms or corridors.}
% \ycheng{
    % Monocular layout estimation has been widely studied \cite{hirzer2020smarthypotheis, 2020_TransMult_Room_lY, hsiao2019flat2layout_room_ly, Lee_2017_ICCV_room_ly,wang2021led2, Pintore_atlanta_net2020, sun2019horizonnet, sun2021hohonet, zou2018layoutnet} and can robustly estimate the room layout.
    % Monocular layout estimation has been widely studied as a direct solution for room layout geometry~\cite{2020_TransMult_Room_lY, hsiao2019flat2layout_room_ly, Lee_2017_ICCV_room_ly,wang2021led2, Pintore_atlanta_net2020, sun2019horizonnet, sun2021hohonet}.
    % % hirzer2020smarthypotheis, zou2018layoutnet
    % However, even by leveraging large field-of-view images, i.e., 360-images, some issues remain unsolved. For example, the room layout estimation may differ with different camera positions, or it may fail to capture large rooms or corridors.
    % On the other hand, reconstructing the floor plan using multiple room layouts is challenging because of the noise caused by camera positions and the scale difference between visual odometry and layout.
%     \TODO{Furthermore, to reconstruct the floor plan using multiple room layouts, the scale difference between estimated camera poses and layouts is an issue in the context of a monocular framework.
% % , different camera location along the whole scene is required, which
%     }
\kike{Furthermore, in the context of a monocular framework, the scale difference between estimated camera poses and the layout projection is a challenge for reconstructing the floor plan using multiple layout estimations directly.}

In this paper, we present a direct floor plan estimation algorithm (360-DFPE) that does not rely on active sensors but directly takes a sequence of 360-images as input.
% \justin{
% Different from previous floor plan estimation approaches, we value the ability of sequential reconstruction as an application.
\kike{In contrast to previous approaches}, we value the ability of sequential reconstruction as an application.
% }
% Toward this goal,
Therefore, we design a novel pipeline that registers multiple 360-layout geometries frame-by-frame and reconstructs the floor plan room-by-room. In Fig.~\ref{fig:teaser}, we exemplify our solution with estimated 360-layouts and camera poses at two timestamps.
% 
% \ycheng{
% Our solution leverages a VSLAM system~\cite{sumikura2019openvslam} and a 360-layout estimation algorithm~\cite{sun2019horizonnet} to estimate camera poses and 360-layout geometries, using both off-the-shelf implementations.
% To handle the scale difference between estimated camera poses and 360-layouts, we devise a warm-up stage at the beginning of our system, which uses few initial keyframes to recover the unknown relative scale through a novel entropy minimization formulation. In the sequential pipeline of 360-DFPE, we propose monocular 360-layout registration which projects and aligns the layout geometry of the current keyframe into the world reference. Next, we identify which room the camera belongs through our novel room identification procedure. Moreover, we apply plane estimation using RANSAC~\cite{RANSAC_fischler1981random} to further remove the invalid layout boundaries.
% After a room is completed, we optimize and vectorize the final room shape based on the geometric evidence.
% }

\ycheng{Our solution leverages a VSLAM system~\cite{sumikura2019openvslam} and a 360-layout estimation algorithm~\cite{sun2019horizonnet} to estimate camera poses and 360-layout geometries, using both off-the-shelf implementations.
To handle the scale difference between estimated camera poses and 360-layouts, we use a few initial keyframes to recover the unknown relative scale at the beginning of our system. Additionally, we propose a novel sequential procedure that identifies room instances without knowing the entire scene in advance. Lastly, we filter out the invalid layout geometries and optimize the final room shape in an iterative coarse-to-fine fashion.
% \st{, achieving faster estimation and higher precision.}
% \TODO{The pipeline of our novel sequential floor plan estimation system comprises 360-layout registration, room identification, plane estimation, and room shape optimization.}
}
% 
% \TODO{In essence, the pipeline of 360-DFPE performs sequentially }.
%
% \ycheng{
%     Given an input keyframe with an estimated camera pose, . Then, . Based on the property of layout geometry, we design the  . Moreover, to recover the unknown the visual odometry scale relative to the layout geometry, we directly regress the value formulate an entropy minimization problem.
% }
%
% propose an entropy optimization method that directly regresses the visual odometry scale respect to layout geometry to register and align multiple 360-layout geometries. We design a sequential room identification procedure that identifies when the current camera keyframe leaves, gets inside, or returns to each room.
%
% Note that since this formulation does not have the information of the whole scene in advance, it additionally requires identifying when the current camera keyframe leaves, gets inside, or returns to every room.
%

\kike{Since existing floor plan estimation datasets~\cite{liu2018floornet,chen2019floorSP} do not provide
sequential frame information, we collect a new dataset, MP3D-FPE, using the Matterport3D dataset~\cite{chang2017matterport3d} rendered by the Minos~\cite{savva2017minos} framework.}
% 
% \ycheng{Since existing floor plan estimation datasets ~\cite{liu2018floornet,chen2019floorSP} \kike{do not provide %continuous and 
% the sequential frame information, we collect a new dataset,}
%  \justin{MP3D-FPE}, which contains 50 complex scenes and 687 rooms in total from Matterport3D~\cite{chang2017matterport3d}, rendered by Minos~\cite{savva2017minos}.
% }
% Our experimental results show favorable performance against current state-of-the-art floor plane estimation solutions, i.e., FloorNet~\cite{liu2018floornet} and Floor-SP~\cite{chen2019floorSP}, despite the fact that our method uses only images as input while the aforementioned methods use registered point clouds of the entire scene captured from active sensors.
Our experimental results show favorable performance against current state-of-the-art floor plan estimation solutions, i.e., FloorNet~\cite{liu2018floornet} and Floor-SP~\cite{chen2019floorSP}, despite the fact that these methods use registered point clouds of the entire scene captured from active sensors.
Our main contribution are the follows:
\begin{itemize}
    \item 
        We propose 360-DFPE which aggregates 360-layout geometries and estimates the floor plan sequentially under a monocular framework.
    \item 
        We propose an entropy optimization method that directly regresses the visual odometry scale relative to layout geometry.
        % We propose an entropy optimization method that regresses the ratio between the visual odometry and layout scale, directly registering 360-layout geometries without feature correspondences, \kike{3D points information}, or any iterative-closest-point regression.
    \item
        % Without knowing the whole scene in advance, detecting individual rooms in the floor plan is challenging. Therefore, 
        % We design a sequential room identification procedure that identifies when the current camera keyframe leaves, gets inside, or returns to each room.
        We design a sequential room identification procedure that tracks the corresponding room for every keyframe using only geometric cues.
        % We design a room identification procedure which defines the set of keyframes that belong to a specific room geometry, allowing us to estimate one room layout at the time, hence, reducing the computational cost regardless of the floor plan size.
    \item
        We propose an iterative coarse-to-fine Shortest Path Algorithm (iSPA) to estimate the room shape, which advances previous formulations~\cite{chen2019floorSP, cabral2014piecewise} with faster speed and better accuracy.
    \item Our experimental results on our newly collected MP3D-FPE dataset show that our monocular 360-DFPE outperforms state-of-the-art methods that require the point cloud of the entire scene as input.
\end{itemize}
\section{Related Work}
% \subsection{Room-layout estimation works.}

\begin{figure*}[t!]

\centering
\includegraphics[width=0.9\linewidth]{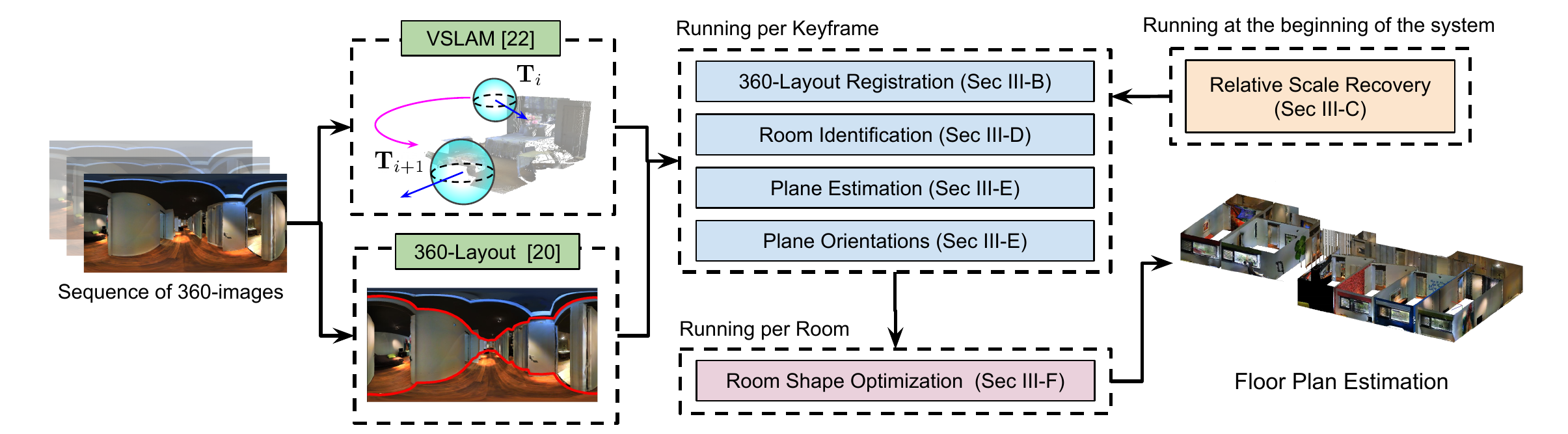}
\vspace{-3mm}
\caption{
% \TODO{
    \textbf{System Pipeline}. Given a sequence of monocular 360-images, our pipeline starts with a visual SLAM and an 360-layout estimation using off-the-shelf implementations. We use the initial few keyframes to recover the unknown relative scale  (Section~\ref{sec:Scale Recovering}). Then, for every keyframe, we sequentially apply 360-layout registration (Section~\ref{sec:360 Registration}), room identification (Section~\ref{sec:Room ID}), plane estimation and plane orientations filtering (Section~\ref{sec:Plane Estimation}).
    Once we detect that a room is completed, we apply the room shape optimization (Section~\ref{sec:Room-Shape Opt}) to estimate the room shape. 
    % \TODO{Except for the scale recovery}, the whole system is able to run in a sequential manner.
% }
}

%%%%
% Input
% SLAM and 360-layout
% VO-scale recovery
% Planes
% Room identification
% filtering
% room shape
%%%%
\label{fig:pipeline}
\vspace{-4mm}
\end{figure*}

\subsection{Single Room Layout Estimation}
\TODO{
    Early works in room layout estimation~\cite{chao2013layout, flint2010_1, flint2011_2, tsai2011real} mainly leverage hand-engineering features on images (e.g., vanishing points, detected objects, and textured regions) to estimate wall-surfaces and wall-contours. In particular, \cite{flint2010_1} defines the room-layout problem as boundary estimation using dominant surface orientations inferred from estimated vanished points, with a predefined room geometry to identify occlusions. 
    % Then, this approach is revisited in~\cite{flint2011_2}, where additional sensor modalities are used to aggregate spatial geometry reasoning. 
    Revisiting \cite{flint2010_1}, \cite{flint2011_2} uses additional sensor modalities to aggregate spatial geometry reasoning. 
    In \cite{tsai2011real}, multiple room boundary hypotheses are aggregated using a Bayesian formulation to achieve real-time room layout reconstruction. Later, \cite{chao2013layout} further considers other cues in the image, such as people, objects, and their geometry relationship. However, all these approaches are constrained by the narrow FoV of the pinhole cameras and usually rely on the assumption that at least two walls are presented in an image.}

    \kike{
       With panorama images (i.e., 360-images), pioneered works like~\cite{zhang2014panocontext, yang2016efficient, xu2017pano2cad} leverage the whole FoV of a scene by projecting multiple 3D room layout hypotheses, estimating vanishing points~\cite{zhang2014panocontext}, room boundaries~\cite{yang2016efficient}, and semantic pixels~\cite{xu2017pano2cad}. However, this 3D projections usually assume some prior geometry constraints such as the camera height, room shape, or the size of some detected objects.
    }
    
    \TODO{
     Deep learning approaches like \cite{dasgupta2016delay_room_LY, Lee_2017_ICCV_room_ly, mallya2015learning, hirzer2020smarthypotheis} robustly estimate room-layout geometry by defining the underlying problem as semantic segmentation, corner junction, or edge map estimation. However, they require a predefined topology to infer the final room-shape structure.}
    Current state-of-the-art solutions~\cite{sun2019horizonnet, sun2021hohonet, wang2021led2, Pintore_atlanta_net2020} directly estimate the room layout boundary from a single 360-image. For instance, HorizonNet\cite{sun2019horizonnet} and HoHoNet~\cite{sun2021hohonet} regress the ceiling-wall, and floor-wall boundaries, assuming that a vertically aligned image can be encoded as a sequence of image-column data. 
    %     % Meanwhile, works like Dula-Net~\cite{yang2019dula}, AtlantaNet~\cite{Pintore_atlanta_net2020} and LED$^2$-Net\cite{wang2021led2} additionally constraint the problem by  the 2D geometry-aware information
    %     % considering into their formulation the room shape and the camera-to-boundary distance. 
    On the other hand, AtlantaNet~\cite{Pintore_atlanta_net2020}, and LED$^2$-Net\cite{wang2021led2},  additionally constraint the problem by adding information from the layout projected into a 2D map.
    % \kike{
    %     Nevertheless, all aforementioned solutions can only project room-layout into 3D by assuming a known geometry constraint usually defined by the camera height. This forces to project the estimated room-layout up to an unknown scale, hereinafter referred as the layout scale. 
    % }
% \kike{
%     In in work, we leverage HorizonNet~\cite{sun2019horizonnet} as the backbone solution to estimate 360-layout geometries registered by SLAM-based camera poses directly. We argue that this is minimal and sufficient information to infer the floor plan geometry.
% }
    % 
% \subsection{Floor Plan Estimation works}
\subsection{Floor Plan Estimation}
    \kike{
        Estimating a floor plan geometry can only be accomplished by using multiple measurements at different scene locations. To this end, early works like~\cite{cabral2014piecewise, pintore20183d} leverage a standard SfM, MVS, and pixel semantic segmentation to project labeled 3D points into a 2D map. Upon this 2D projection, \cite{cabral2014piecewise} formulates the floor plan reconstruction as a shortest path problem, and~\cite{pintore20183d} formulates it as non-linear optimization, using projected floor and ceiling pixels as internal space of the room, and wall pixels as a room boundary evidence.
        % For projecting floor and ceiling planes. These solutions infer the camera height from 3D-projected sparse points \cite{cabral2014piecewise} assuming a Manhattan constraint~\cite{flint2011_2}.
        However, these methods rely on the quality and sparsity of the projected point cloud, which may vary for scenes with poor textures. 
        % Additionally, both methods requires the entire point cloud scene to reason about individuals rooms; hence a sequential estimation is not possible. 
    }
    % \justin{
    %     % \kike{On the other hand, state-of-the-art solutions for floor plan estimation~\cite{wijmans2017exploiting, lin2019floorplanPartialRGBD, le2017denseFPE, liu2018floornet, chen2019floorSP, xiao2014reconstructing} mainly rely on active sensors to acquire 3D information (e.g., depth cameras, laser scanners). For instance, early solutions~\cite{wijmans2017exploiting, le2017denseFPE} leverage RGB-D images to build a 3D scene from where the floor plan structure can be inferred using normal surfaces . However these approaches rely on semantic estimations over the scene to correctly identify room geometries, and doors. Solutions like \cite{xiao2014reconstructing} }
    %     On the other branch, some methods \cite{wijmans2017exploiting, zhang2013estimating, lin2019floorplanPartialRGBD, le2017denseFPE, liu2018floornet, chen2019floorSP, xiao2014reconstructing}
    %     estimate the floor plan using 3D point cloud as input or the density map, the 2D projection of the point cloud. These methods rely on active sensors such as depth cameras or laser scanners that require extensive pre-processing to reconstruct the point cloud.
    %     Among them, FloorNet~\cite{liu2018floornet} estimates low-level geometry with multiple network branches and aggregates them with integer programming. Floor-SP~\cite{chen2019floorSP} formulates the shortest path algorithm (SPA) with several constraints following~\cite{cabral2014piecewise} to optimize the room shape based on the corner, edge, room prediction using deep neural networks.
    % }
    
    \kike{
        % Without any doubt, the 
        The most stable and robust solutions for floor plan estimation~\cite{xiao2014reconstructing, lin2019floorplanPartialRGBD, liu2018floornet, chen2019floorSP, stekovic2021montefloor} mainly rely on active sensors (e.g., LiDAR or depth cameras) to directly acquire 3D information of the scene. For instance, early works such as~\cite{xiao2014reconstructing} leverage multiple 3D LiDAR-based scans by estimating primitive geometries from a layered-up formulation. In contrast, works like~\cite{lin2019floorplanPartialRGBD, liu2018floornet} define the problem by acquiring 3D data from dense depth images. Floorplan-Jigsaw~\cite{lin2019floorplanPartialRGBD} constraints the problem as a set of data with small overlapping, searching for the most coherent alignment in 2D; FloorNet~\cite{liu2018floornet} estimates low-level geometry with multiple Convolutional Neural Network branches and aggregates the outputs with integer programming, assuming a predefined room-shape.
    }
    
    \justin{
        Current state-of-the-art solutions~\cite{chen2019floorSP, stekovic2021montefloor} assume that the whole scene has been scanned and registered into a 3D point cloud as input, which is later projected into a 2D density map. From this 2D map, Floor-SP~\cite{chen2019floorSP} views the floor plan reconstruction as multiple single room estimations, identifying room instances with Mask R-CNN~\cite{he2017mask}, then formulating the room geometry estimation as the shortest path problem similar to~\cite{cabral2014piecewise}. Recently, MonteFloor~\cite{stekovic2021montefloor} adopts the pipeline similar to Floor-SP and introduce Monte Carlo Tree Search (MCTS) for searching and evaluating estimated room candidates.
    }

\section{Method}
% \kike{
%     We present 360-DFPE, a direct floor plan estimation algorithm that takes a sequence of 360-images as an unique input and estimates a floor plan geometry sequentially, i.e., without collecting the whole scene data in advance.
%     % Our approach leverages out-off-shelf solutions for 360-room layout estimation~\cite{sun2019horizonnet, sun2021hohonet} and the monocular VSLAM pipeline~\cite{sumikura2019openvslam}, projecting 360-layout geometries and estimating camera poses in the scene. 
%     For illustration purpose, our pipeline is presented in Fig.~\ref{fig:pipeline}, and our general overview is as follows.
%     }

\subsection{Overview}
    % We present 360-DFPE, a direct floor plan estimation algorithm that takes a sequence of 360-images as a unique input and estimates a floor plan geometry sequentially, i.e., without collecting the whole scene data in advance. For illustration purposes, our pipeline is presented in Fig.~\ref{fig:pipeline}.

\kike{ 
    We present 360-DFPE, a direct floor plan estimation algorithm that takes a sequence of 360-images as input to estimate a floor plan geometry sequentially. For illustration purposes, our pipeline is presented in Fig.~\ref{fig:pipeline}, where our proposed solution starts with estimating cameras poses and 360-layouts using ~\cite{sumikura2019openvslam, sun2019horizonnet} with a loosely-coupled integration.}

\kike{
    Since both mentioned methods are monocular solutions, their estimations are defined up to different and unknown scales. Therefore, at the beginning of the system, a relative scale recovery estimates the missing odometry scale with respect to the layout geometries (see Section~\ref{sec:Scale Recovering}).
}

\kike{
    Then, running every keyframe, the estimated layout is aligned using VSLAM-based camera poses by a 360-Layout Registration (see Section~\ref{sec:360 Registration}). Next, to identify 
    % and reconstruct 
    every room in the scene, an identification algorithm tracks the current camera location sequentially (see Section~\ref{sec:Room ID}). In addition, each estimated layout is decomposed into a set of plane geometries to remove noisy estimations. Upon these plane geometries, we also compute the plane orientations of each room~(see Section~\ref{sec:Plane Estimation}).
    % Lastly, towards reconstructing every room as a set of corners, a room shape optimization procedure is executed to estimate the room corners once  the camera leaves the room (see Section~\ref{sec:Room-Shape Opt}).
    Lastly, once the camera leaves each room, a room shape optimization procedure estimates the room geometry as a set of corners, which vectorizes each room and the whole floor plan structure~(see Section~\ref{sec:Room-Shape Opt}).}
\subsection{360-Layout Registration}
\label{sec:360 Registration}
\kike{
        % The underlying problem in this section is to define a geometry registration that uses VSLAM-based camera poses and 360-layout estimations, where both are defined under unknown and different scales.
        \ycheng{
        The underlying problem in this section is to register multiple 360-layout estimations using VSLAM-based camera poses, \kike{where both} are defined under unknown and different scales.
        }
        % The underlying problem in this section is to define a parametrized geometry registration using 
        % VSLAM-based camera poses and 360-layout estimations, where both are defined under unknown and different scales. For this purpose, we first define a 360-layout as a set of projected 3D points $\mathbf{x}$ in the Euclidean space, using the estimated floor-wall boundaries \dennis{again, the definition of boundaries is not clear} from HorizonNet~\cite{sun2019horizonnet}, as follows:
        % For this purpose, we first define the layout geometry as a set of projected 3D points $\mathbf{x}$ in the Euclidean space, using the layout boundaries estimated by HorizonNet~\cite{sun2019horizonnet}. This projection is described as follows:
        % 
        \kike{For this purpose, we first define a layout boundary as a set of 3D points, which are projected from a 360-layout estimation~\cite{sun2019horizonnet}, defined by $W$ points on an equirectangular image with size $H\times W$. A point $\mathbf{x}$ of this layout boundary is defined as follows:}
        % 
        % \ycheng{For this purpose, we first define the layout boundary as a set of projected 3D points $\mathbf{x}$ in the Euclidean space from the 360-layout estimated by HorizonNet~\cite{sun2019horizonnet}.
        % Given an equirectangular image with size $H\times W$, HorizonNet outputs $W$ points. We project them into 3D as follows:
        % }
        % \dennis{again, the definition of boundaries is not clear}
        % \kike{For this purpose, we first define the layout boundary of the $i$-th keyframe by $\mathbf{P}_i=[\mathbf{x}_1, \cdots, \mathbf{x}_{W}] \in \mathbb{R}^{3\times W}$ projected from a 360-layout estimation~\cite{sun2019horizonnet}, defined on an image with size $H\times W$. A point $\mathbf{x}$ in this layout boundary is defined as follows:}
}
% \kike{
%         The underlying problem in this section is to coherently register multiple 360-layout estimations using VSLAM-based camera poses, where both the layout projection and the odometry estimation are defined under unknown and different scales. For this purpose, we first define a projected 360-layout as a set of 3D points $\mathbf{x}$ in Euclidean space, using the estimated floor-wall boundaries, as follows:
% }
\begin{equation}\label{eq:xyz-def}
    \mathbf{x} =
    \begin{bmatrix}
        x \\ y \\ z
    \end{bmatrix}
    =
    \begin{bmatrix}
        \frac{h}{\sin\phi} \cos\phi \sin\theta \\
        -h \\
        \frac{h}{\sin\phi} \cos\phi \cos\theta
    \end{bmatrix},
\end{equation}
\kike{
where $h$ is the current camera height (usually assumed or measured in advance~\cite{pintore20183d, cabral2014piecewise, sun2019horizonnet, wang2021led2}), and $(\theta, \phi)$ is its \ycheng{spherical coordinate on the equirectangular image.}
}
% For simplicity, we assume that the estimated layout boundary on the equirectangular image consists of $W$ points. 
% \dennis{why 1024 points?}.
% \ycheng{We represent the 3D layout boundary of the $i$-th keyframe by}

\kike{
    \ycheng{We represent the layout boundary of the $i$-th keyframe as the set of points}
    % Therefore, we can represent the $i$-th projected layout boundary in 3D by
    $\mathbf{P}_i=\{\mathbf{x}_1, \cdots, \mathbf{x}_{W}\}$. 
    Additionally, we define the first estimated camera pose as the world coordinate references and set its distance to the floor \justin{(i.e., camera height)} to an arbitrary constant value, i.e., $h=1$, which defines the layout projection scale. 
    % \justin{Without the true camera height,% the real scale of 360-layout projection is unknown.}
    % {a 360-layout projection is defined only under an unknown scale, hereinafter referred as the layout scale.}
} 

\begin{figure*}[t!]

% \TODO{Put text under (a), (b), (c)}
\centering
\includegraphics[width=0.9\linewidth]{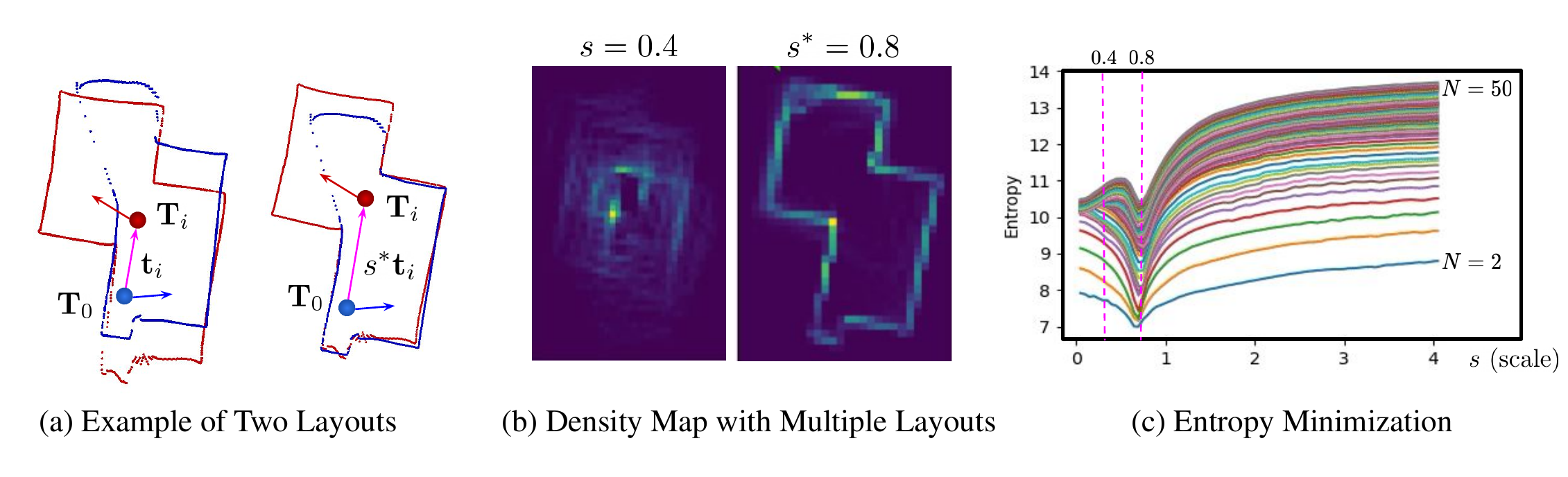}
\vspace{-6mm}
\caption{
\textbf{360-Layout Registration with Relative Scale Recovery.}
% \ycheng{
%     (a) We illustrate the issue of missing relative scale between visual odometry and the layout. Without recovering the $S_{VO}$ defined in Eq.~\ref{eq:registration}, the layouts estimated in the initial camera pose $C_0$ (blue) and another camera pose $C_i$ (red) cannot be aligned properly.
%     }
\kike{
    (a) We illustrate registering two 360-layout boundaries (red and blue lines) from layout estimations by \eqref{eq:registration}. By scaling the relative translation $\mathbf{t}_i$ with $s^*$, we can align them properly.
    }
% In the left, given two layouts estimated in the initial camera pose $C_0$ and another camera pose $C_i$, the layouts cannot be aligned properly because the relative scale difference . Therefore, we need to recover the relative scale in Eq.~\ref{eq:registration} as shown in the right.
% \dennis{Notations are not consistent with the text and I changed them. Btw, what is $t^W_i$ as we do not have it in the text.}
% In Panel (a), two room layouts are presented i.e., red and blue boundaries. For illustration purposes, two corresponded points from these boundaries are also presented, i.e., $\mathbf{p}^i$ and $\mathbf{p}^{i-1}$.
% Note that both points are defined at $C^i$ and $C^{i-1}$ camera references from which, in turn, they are registered in world coordinates by the camera poses $T^W_i$ and $T^W_{i-1}$. Therefore, by optimizing over $T^W_i$ and $T^W_{i-1}$, we can align both boundaries, see Sec.~\ref{sec:Scale Recovering}. 
% \ycheng{(b) To recover the scale, we project multiple layouts into a 2D probability density map using Eq.~\ref{eq_pdf} and minimize the entropy.
% The left and the right shows the visualization of the density map under different $S_{VO}$ values. It can be observed that when $S_{VO}=0.8$, the layouts are aligned more correctly; it will thus yield a lower entropy value for the probability density map.}
\kike{
    (b) We visualize multiple layout estimations with 2D density maps using different $s$ values, e.g., 0.4 and 0.8. The density map using $s^*=0.8$ will yield a lower entropy due to better alignment.
    % In (b), we project multiple layout estimations into two 2D density maps using \eqref{eq_pdf_registration} with different $S_{VO}$ values, e.g., 0.4 and 0.8. Note that the density map that uses $S^*_{VO}=0.8$ shows less disorder which will yield a lower entropy than setting $S_{VO}=0.4$.
}
\kike{
    (c) We plot the entropy values under different scale $s$ using different number of layouts $N$. Note that we can find the optimal relative scale value for the scene through entropy minimization. See Section~\ref{sec:Scale Recovering} for more details.
    % In (c), we present several entropy evaluations using different $N$ numbers of slide-windows. Note that despite the minimum entropy varies for every evaluation, a clear minimum can be identified at $S^*_{VO}=0.8$ representing the optimal relative scale value for the scene. For more details cf. Section~\ref{sec:Scale Recovering}.
}
}

% \ycheng{
% (c) We show the plot of entropy defined in Eq.~\ref{eq_scale_rec} using the same example, the optimal relative scale $S_{VO}^*$ for the minimal entropy is $0.8$.
% \dennis{It seems some details are not in the main text. Captions should only reflect some additional explanations, on top of of self-contained explanations of the main text.}
% }}
% \kike{

% }
\label{fig:registartion}
\vspace{-4mm}
\end{figure*}

\kike{
    To define the registration of $\mathbf{P}_i$ into the world coordinate, we leverage its VSLAM-based camera pose as follows:}
\begin{equation}\label{eq:registration}
    \mathbf{P}^W_i =\{\mathbf{R}_i \cdot \mathbf{x} + s ~\mathbf{t}_i~|~ \forall \mathbf{x} \in \mathbf{P}_i\},
\end{equation}
\kike{
    where $\mathbf{T}_i =[\mathbf{R}_i| \mathbf{t}_i] \in SE(3)$ is the estimated camera pose with rotation and translation, and $s$ is the unknown visual odometry scale for a monocular VSLAM solution~\cite{scaramuzza2011visual, sumikura2019openvslam}. This visual odometry scale and the layout projection scale are not necessarily equal.}
% \kike{For reference purpose, in Fig.~\ref{fig:registartion}-(a), we exemplify the registration of two layout boundaries, where by }
% 
\kike{
    Note that by changing \ycheng{the scale} $s$ in \eqref{eq:registration}, we can modify the translation of $\mathbf{P}_i$ in the world coordinate. For reference purpose, in Fig.~\ref{fig:registartion}-(a), the registration of two layout boundaries are illustrated.
    % This is our 360-layout registration.
    % In Fig.~\ref{fig:registartion}-(a), we exemplify the registration of two layout boundaries.
}

\subsection{Relative Scale Recovery}
\label{sec:Scale Recovering}
% \justin{
    % Without loss of generality,
    % We define the registration of $N$ projected layouts as a function of $s$ by concatenating $\mathbf{P}^W_i$ for $i \in [0, N-1]$ into $\hat{\mathbf{P}}^W \in \mathbb{R}^{3\times WN}$, hence formulating an optimization problem as follows:
    % ected $\hat{\mathbf{P}}^W$ boundaries. In Fig~\ref{fig:registartion}-(a), we illustrate the registration of two layout boundaries, showing intuitively how different $S_{VO}$ values defines different layout registrations.  
% }

\ycheng{
    \kike{To recover the unknown visual odometry scale $s$ relative to the 360-layout estimates,}
    % To recover the unknown relative scale between the visual odometry and 360-layouts, 
    we aggregate a sequence of registered layout boundaries and optimize over $s$,
    constrained under our predefined layout projection scale (i.e., $h=1$, see Section~\ref{sec:360 Registration}).% \TODO{assuming the 360-layout scale is $1$.}
}
\ycheng{
    We formulate an optimization problem with $N$ layout boundaries as a function of $s$:
}

    % Note that under the monocular setup, visual odometry scale is defined by the first frames \justin{in} VSLAM initialization \cite{scaramuzza2011visual, sumikura2019openvslam}. However, the relative scale, which is denoted as $S_{VO}$, with respect to the layout scale remains unknown. Therefore, we formulate an entropy minimization problem as a function of $S_{VO}$ to recover the relative scale. First, we concatenate $\mathbf{P}^W_i$ for $i=[0,N]$ into $\hat{\mathbf{P}}^W \in \mathbb{R}^{3\times 1024N}$. This optimization is presented as follows:

\vspace{-3mm}
\begin{equation} \label{eq_pdf_registration}
% \rho_{u, v}(S_{VO}) = \Phi (
%     \hat{\mathbf{P}}^W 
% )_{[u, v]}
F_{s} = \Phi (
\mathbf{P}^W_0 \cup \cdots \cup \mathbf{P}^W_{N-1}
    % \hat{\mathbf{P}}^W 
)
% _{[u, v]}
\end{equation}
% \begin{equation} \label{eq_scale_rec}
% S^*_{VO}= \argmin_{S_{VO}}\sum_{u, v}-\mathcal{F}_{u,v}(S_{VO})({u,v})\cdot \log~\mathcal{F}_{u,v}(S_{VO}),
% \end{equation}
\begin{equation} \label{eq_scale_rec}
s^*= \argmin_{s}\sum_{u, v}-F_{s}(u, v)\cdot \log{F_{s}(u, v)},
\end{equation}
\kike{
    where $\Phi(*)$ is a top-down projection function that computes a 2D normalized histogram for all the input points, defining the 2D probabilistic density map $F_{s}$, and \eqref{eq_scale_rec} defines an entropy minimization for $s$.
    % $\mathcal{F}_{u, v}$ as a function of $S_{VO}$.
}
\ycheng{
    %  \justin{
        %  where $\Phi(*)$ is a top-down projection function that maps \kike{$\hat{\mathbf{P}}^W$} into a 2D probabilistic density map represented by $\rho_{u, v}$.}
        %  \dennis{Not clear how this $\Phi(*)$ function is calculated.}
        The intuition is that only when $s$ is correctly estimated, the layout boundaries in multiple consecutive time frames can be well-aligned on the 2D density map, which will yield \justin{the lowest} entropy value.
        %  We provide an example of the probability density map and entropy in Fig.~\ref{fig:registartion}~(b-c) over different $S_{VO}$.
        % %  For illustration purposes see Fig.~\ref{fig:registartion}-(a)(b).
        %  \dennis{Looks like there are some issues for the reference in Fig. xx (also elsewhere), i.e., it should be Fig. 3 instead of Fig. III here.}
        %  \dennis{the caption of Fig. 3 (b) is not mentioned above in the text. This is confusing to refer it here.}
        }
    %  \justin{Since $S_{VO}$ is a scalar, we apply linear search to obtain $S_{VO}^*$ that minimizes the total entropy in every coordinates.}
  
    \kike{
    To solve~\eqref{eq_scale_rec}, we apply a linear search in a predefined range as studied in~\cite{nocedal2006numerical}. 
    \ycheng{
        Ideally, using all the keyframes will bring the optimal $s^*$ for the scene.
        However, considering the trade-off between data usage and performance, we devise a warm-up stage in the beginning of the system that takes the initial 20\% keyframes to recover a stable $s^*$.
    }
    We use a 3-level coarse-to-fine search with $\{0.5, 0.1, 0.01\}$ as the step size and a sliding-window with size $N=10$.
    }
    % TODO{This allows us to compute and update $S^*_{VO}$ at every new keyframe.}
    % For more details see Fig.~\ref{fig:registartion}.% }

\kike{
    % In Figure~\ref{fig:registartion}-(a), we exemplify how a different $S_{VO}$ value can align two 360-layout geometries by scaling their camera poses; 
    In Fig.~\ref{fig:registartion}-(b), we illustrate the density function $F_{s}$, showing how $s^*$ defines a better alignment of multiple layout geometries; Fig.~\ref{fig:registartion}-(c) shows that the minimum entropy value is at the optimal $s^*$ scale under different sliding-window sizes.
    }
\subsection{Room Identification}
\label{sec:Room ID}
% \TODO{
% \begin{itemize}
%     \item A room Identification is the result of evaluating a likelihood value on a density function $\mathcal{H}$.
%     \item This 2D density function is build by sequentially aggregating multiple patches $\mathbf{Q}_i$.
%     \item Every patch is defined as a close region around each camera poses, bounded by a constant distance d or by the $\mathbf{P}_i$ defined in Sec.~\ref{sec:Scale Recovering} (intersection)
% \end{itemize}
% }

\begin{algorithm}[tb]
\caption{Room Identification}\label{alg:room_id}
\begin{algorithmic}[1]
\renewcommand{\algorithmicrequire}{\textbf{Input:}}
\renewcommand{\algorithmicensure}{\textbf{Output:}}
\Require New camera pose $\mathbf{T}_i = [\mathbf{R}_i| \mathbf{t}_i]$ and  a set of room density maps $\mathbf{H} = \{ H_j \}_{j=1}^m$
% \Ensure $\{ \mathcal{H}_j \}_{j=1}^M$   \Comment{$M$ room density maps}
% \State $M \gets 0$
% \For{$i=1$ to $N$}
% \State Compute $\mathbf{B}_i$ by
\State $\mathbf{Q}_i \gets$ Bounded 2D area enclosed by $\mathbf{B}_i$ using \eqref{eq_patch}
\State $(u_i, v_i) \gets$ 2D projected coordinate of $s^*~\mathbf{t}_i$

\If{$H_m(u_i,v_i) \geq 0.5$}
    \State $H_m \gets H_m + \Phi(\mathbf{Q}_i)$ \Comment{In current room $m$}
    \State $H_m \gets H_m~/~\max_{u, v}H_m(u,v)$
\ElsIf{$\max_{j < m} H_j(u_i,v_i) \geq 0.5$}
    \State $H_j \gets H_j + \Phi(\mathbf{Q}_i)$ \Comment{In previous rooms $j$}
    \State $H_j \gets H_j~/~\max_{u, v}H_j(u,v)$
\Else 
    \State $H_{m+1} \gets \Phi(\mathbf{Q}_i)$ \Comment{Create a new room $m+1$}
    \State $H_{m+1} \gets H_{m+1}~/~\max_{u, v}H_{m+1}(u,v)$
    \State  $\mathbf{H} \gets  \mathbf{H} + \{ H_{m+1} \}$
\EndIf
% \EndFor
% \Require The camera pose of the $i$-th keyframe on 3D $\mathbf{t}_i=[x_i, y_i, z_i]$ and 2D projection $\mathbf{c}_i=[u_i, v_i]$. The current room id $j$ and the set of room 2D density functions $\mathcal{R}=\{ \mathcal{H}_k\}_{k=1}^j$.
% \Ensure The room id of the $i$-th keyframe, which will be in $[1, j+1]$.
% % \State $\mathbf{D}_i \gets \{[u, v]~\mid~\left|[u, v] - [u_i, v_i]\right| \leq d \}$ 
% % \left \|  \mathbf{x} \right \|_2\
% \State $\mathbf{D}_i \gets \{(x, 1, z): \left \|(x, z) - (x_i, z_i) \right \|_2 \leq r \}$ 
%     \Comment{Circle area with radius $r$}
% \State $\mathbf{R}_i \gets$ the bounded 2D area enclosed by the layout $\mathbf{P}^W_i$
% \State $\mathbf{Q}_i \gets \mathbf{D}_i \cap \mathbf{R}_i$
% \State $\mathcal{H}_j \gets \mathcal{H}_j + \Phi(\mathbf{Q}_i)$
% \State $\mathcal{H}_j \gets \mathcal{H}_j~/~\max_{u, v}\mathcal{H}_j(u,v)$ 
%     \Comment{Normalize by the largest value}
% \If{$\mathcal{H}_j(u_i,v_i) \geq 0.5$}
%     \State \Return $j$
%     % \State $N \gets \frac{N}{2}$  \Comment{This is a comment}
% \ElsIf{$\max_{k < j} \mathcal{H}_k(u_i,v_i) \geq  0.5$}
%     \State \Return k
% \Else
%     \State Create and initialize the new room $j+1$
%     \State \Return $j+1$
% \EndIf
\end{algorithmic}
\end{algorithm}

\kike{
%   The reconstruction of the floor plan structure can be decomposed into individual room estimations as studied in~\cite{liu2018floornet, chen2019floorSP, pintore20183d}. Thus, the underlying challenge is to identify whether the current camera pose is inside, leaving, or returning to a room. For this purpose, we propose using a 2D room density function $\mathcal{H}(u, v)$ which evaluates the likelihood of a camera pose belonging in the respective room.
}
\ycheng{
    % The reconstruction of the floor plan structure can be decomposed into individual room estimations as studied in~\cite{liu2018floornet, chen2019floorSP, pintore20183d}.
    In a sequential framework, the entire scene structure is unknown in advance. Therefore, we formulate the problem of identifying room instances as tracking the belonging room for every incoming keyframe.
    % belongs the room where they.
    We propose using the layout geometries to construct a 2D room density function that evaluates the likelihood of a camera position in the room.
    }

\ycheng{
    Our assumption is that the layout boundaries should change smoothly along the camera trajectory unless the camera crosses the room edge. Hence,}
\kike{
   to build the density function $H$, we sequentially aggregate the
   2D area inside the clipped layout boundary $\mathbf{B}_i$ around every camera position. This boundary is defined as follows:}
% \begin{gather}\label{eq_patch}
%     \mathbf{B}_i = \{
%     \rho_r(\mathbf{x} - \mathbf{t}_i) +  \mathbf{t}_i \mid \forall \mathbf{x} \in  \mathbf{P}_i^W \}, \\
%     \rho_r(\mathbf{x}) = 
%     \begin{cases}
%     \mathbf{x} , 
%         & \text{if } \| \mathbf{x}\| \leq r\\
%     r \frac{\mathbf{x}} {\| \mathbf{x} \|}, & \text{otherwise}.
%     \end{cases}
% \end{gather}
\begin{gather}\label{eq_patch}
    \mathbf{B}_i = \{
    \mathbf{R}_i\cdot 
    \rho_r(\mathbf{x}) +  s^* ~\mathbf{t}_i \mid \forall \mathbf{x} \in  \mathbf{P}_i \}, \\
    \rho_r(\mathbf{x}) = 
    \begin{cases}
    \mathbf{x} , 
        & \text{if } \| \mathbf{x}\| \leq r\\
    r \frac{\mathbf{x}} {\| \mathbf{x} \|}, & \text{otherwise}.
    \end{cases}
\end{gather}
\kike{
where
%  $\mathbf{P}^W_i$ is the layout boundary in world coordinates described in Section~\ref{sec:Scale Recovering}, and
$\rho_r(\mathbf{x})$ is a clipping function with a radius $r$
% that defines the boundary $\mathbf{B}_i$ in order 
in order to avoid over-trusting the original layout boundary $\mathbf{P}_i$, especially those regions far from the camera.
}

\ycheng{
    The proposed room identification procedure is detailed in Algorithm~\ref{alg:room_id}. \justin{Given an input keyframe with its camera pose $\mathbf{T}_i$, we first determine if the camera position is inside the current room or whether it is entering one of the previously visited rooms, using the $H$ function of each room.} If one condition is satisfied, we update the belonging room's density function with the area bounded by $\mathbf{B}_i$. Otherwise, we will trigger room creation and initialize a new density map.
}

% \kike{
%     The proposed room identification procedure is detailed in Algorithm~\ref{alg:room_id}, where an input keyframe, registered with the camera pose $C_i$, is evaluated with the current room's density function $\mathcal{H}_m$, if the camera pose is inside the current room $\mathcal{H}_m$ or whether it is entering one of the previously visited rooms $\mathcal{H}_j$, we update the room's $\mathcal{H}$ using the area bounded by $\mathbf{B}_i$. Otherwise, we will trigger creating a new room and initialize the new density map.
% }
    % defining whether the keyframe is inside or outside the current room. In the case of belonging to the room, $\mathcal{H}_m$ is updated. Conversely, the highest likelihood evaluation using previous room density functions $\mathcal{H}_j$ will be selected and updated as the new current room. If none of the existing density rooms evaluate a likelihood greater than a predefined threshold (e.g, 0.5), a new room will be created and initialized with a new density map $\mathcal{H}_{m+1}$.

\begin{figure}[t!]

% \TODO{Horizontal fig}
\centering
\includegraphics[width=0.85\linewidth]{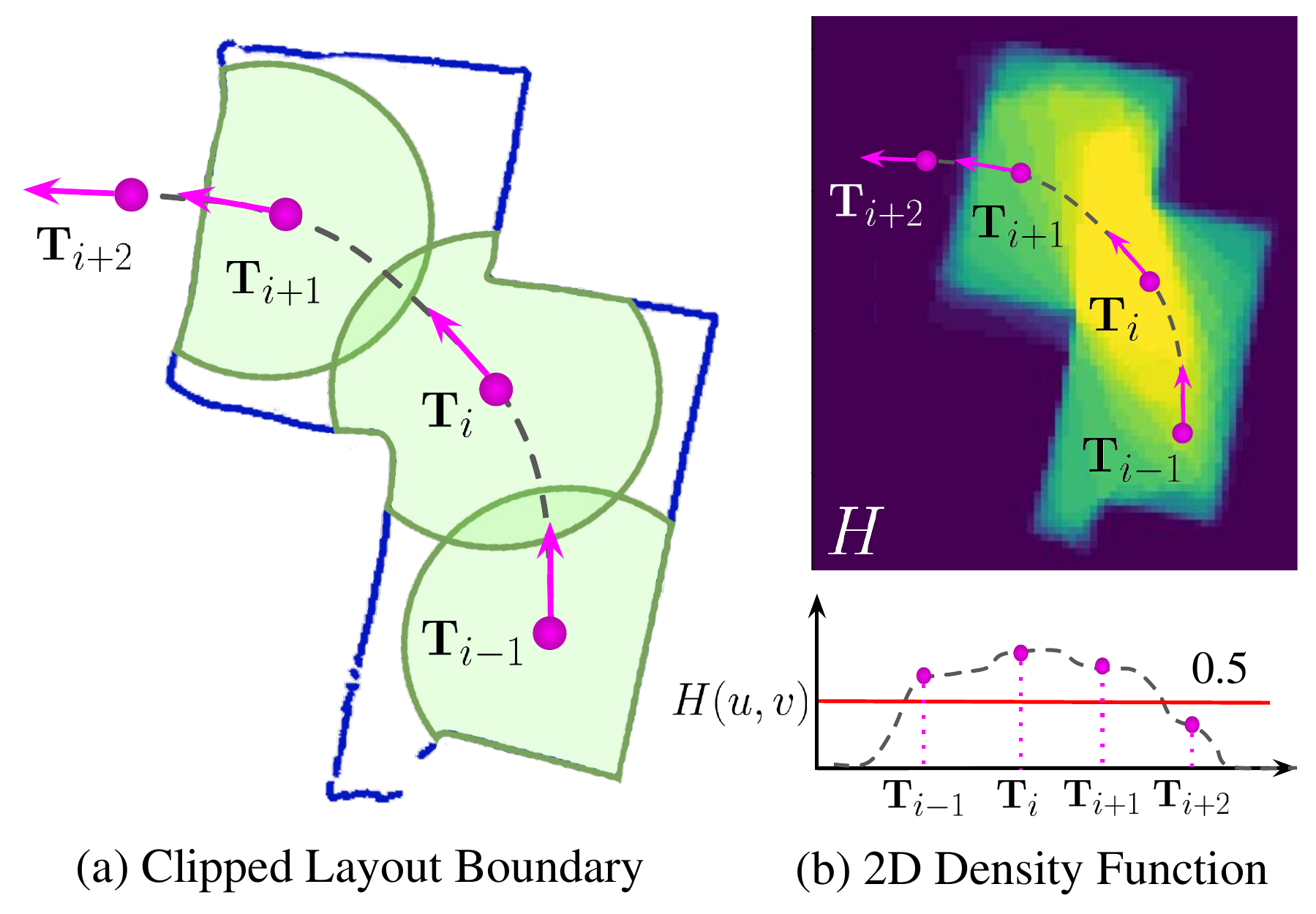}
\vspace{-3mm}
% \fbox{\rule{0pt}{2.5in} \rule{.9\linewidth}{0pt}}
\caption{
\textbf{Room Identification.} (a) The green circles are the clipped layout boundaries (see \eqref{eq_patch}). To track if the camera is inside a room, we sequentially aggregate them and build a 2D density function for the room. (b) If the new camera pose $\mathbf{T}_{i+2}$ has a density value lower than the threshold 0.5, we consider that the camera has crossed the room boundary.
% \ycheng{
%     % \textbf{Room Identification and Room-shape Optimization.}
%     \textbf{(a-b) Room Identification.} To cluster the camera poses and the corresponding layouts that are within the same room, we project the piece-wise planes (orange) and camera poses (pink) onto a 2D grid.
%     If the edge between a new camera pose $C_{n+1}$ and the room center $O_R$ is cross any plane, we consider that the room is completed and group the camera poses within the room.
%     \textbf{(c-d) Room-shape optimization.} Inspired by \cite{chen2019floorSP}, we perform optimization over the room shape represented by corners $\mathbf{p}_i$ based on the information of planes and filtered wall orientations $\hat{\theta}$ of the room by Shortest Path Algorithm (SPA). Note that some of the rooms are not entered to be identified.
% }
}\label{fig:room_id}
\vspace{-4mm}
\end{figure}

For illustration purpose, in Fig.~\ref{fig:room_id}-(a), we present a sequence of keyframes with their corresponding clipped layout boundary $\mathbf{B}_i$. In Fig.~\ref{fig:room_id}-(b), we depict the density value $H(u, v)$ for several keyframes. Note that the keyframe at $\mathbf{T}_{i+2}$ is identified outside the room since its evaluation in $H(u, v)$ is lower than a predefined threshold.
% \TODO{In our experiments, we  empirically set the threshold equal to 0.5.}
\TODO{In our experiments, we find that setting this threshold equal to 0.5 strikes a good balance between false negatives and false positives for room identification.}
% \TODO{In our experiments, we  empirically set this threshold equals to 0.5 balancing between false negative and false positive room identifications.}
    % In our implementation, this threshold have been set equal to 0.5 empirically.
    % , which ensures that at least half of the aggregated data defines the density function $H$ for every room.
    % \dennis{Add the meaning of the score lower than 0.5. \kike{(done: is it clear?)}}

\vspace{-2mm}

\subsection{Plane Estimation from Layout Geometry}
\label{sec:Plane Estimation}
% \TODO{
% \begin{itemize}
%     \item Every 360-Layout from every keyframe is decomposed into plane-wall geometries using RANSAC.
%     \item A plane feature is represented by:
%     \begin{equation}\label{eq_planes}
%         \mathbf{l}_r(\mathbf{n}_r, d_r) = \left\{\mathbf{p}:~~ d_r = \mathbf{p} \cdot \mathbf{n}_r~\forall~\mathbf{p} \in \mathbf{P}^{O_i} \right\},
%     \end{equation}
%     \item Constrast the RANSAC with the use of Hough transform in Pintore's work
%     \item 
%     \begin{equation}\label{eq_bayesian_filter}
% \begin{aligned}
%     &p(\hat{\theta}_{r}) = \eta \cdot p(\hat{\theta}_{r}) \mathcal{N}(\theta_r, \sigma_{\theta_r}),\\
%     &\sigma_{\theta_r} = \sigma_{\theta} + \lambda_\theta d^i_r~,
% \end{aligned}
% \end{equation}
% \end{itemize}
% }
\kike{
    Considering that a 360-layout estimation may contain noise and even invalid geometries,
    % we decompose every estimated boundary $\mathbf{P}_i$ (see Section~\ref{sec:Scale Recovering}) into subsets of points, i.e., $\mathbf{S}_0\cup\cdots\cup\mathbf{S}_k = \mathbf{P}_i$, and remove the undesired subsets.
    we decompose every set $\mathbf{P}_i$ (i.e, a layout boundary, see Section~\ref{sec:Scale Recovering}) into subsets of points~$\mathbf{S}_0\cup\cdots\cup\mathbf{S}_k=\mathbf{P}_i$ to identify and remove the undesired geometry regions.
    % where every subset $\mathbf{S}_k$ can be defined by dividing uniformly $\mathbf{P}_i$. 
    In our implementation, we directly leverage the estimated wall corners from~\cite{sun2019horizonnet}, splitting $\mathbf{P}_i$ into several subsets such that each $\mathbf{S}_k$ represents a wall, namely the plane-wall feature.
    }
% \kike{
%     Upon that, we can evaluate them to represent a plane geometry, applying RANSAC~\cite{RANSAC_fischler1981random}; hence pruning out unreliable geometries. Every subset $\mathbf{S}_k$ can be defined by dividing uniformly the estimated boundary $\mathbf{P}_i$. In our implementation, we directly define $\mathbf{S}_k$ by a pair of consecutive corners that are handy estimated from~\cite{sun2019horizonnet}.}

\kike{
    For every plane-wall feature, we formulate a plane estimation problem by using RANSAC~\cite{RANSAC_fischler1981random}. To this end, we sample points from $\mathbf{S}_k$ to compute a plane geometry $(\mathbf{n}, d)$, where $\mathbf{n}$ is the normal vector of the plane, and $d$ is the distance value measured from the camera to the plane.}
% \kike{Then, to remove undesired            geometries
% }
\kike{
    Then, every estimated plane is evaluated up to a maximum residual error equal to $3\times 10^{-2}$, where every subset $\mathbf{S}_k$ that describes an inliers ratio lower than 0.9 is removed as a reliable geometry evidence. These hyper-parameters were chosen empirically.}
% \dennis{Explain a bit the meaning of the inlier ratio.\kike{(done?)}}
    % To evaluate each plane-wall feature, we define a plane estimation problem using RANSAC~\cite{RANSAC_fischler1981random}. For every $\mathbf{S}_k$, we sample points and compute a plane geometry model $(\mathbf{n}, d)$, where $\mathbf{n}$ is the normal vector of the plane, and $d$ is the distance value (i.e., the distance from the camera to the plane).
    % Our RANSAC configuration uses a maximum number iterations equal to 112; hence expecting an outlier ratio of 0.8 with a probability of success around 0.99. 
    % To assert that a plane-wall feature is acceptable as geometry evidence, we reject plane definitions that evaluate an inliers ratio lower than 0.9.
    % \justin{We eliminate plane-wall features with an inlier ratio lower than 0.9 to ensure the quality of the geometry evidence.
    % }

\TODO{
In addition, we estimate the likely plane orientations for each room with Bayesian filtering, which is later used in the room shape optimization as a geometry constraint (see Section \ref{sec:Room-Shape Opt}).} This filtering formulation is presented as:

\begin{equation}\label{eq_bayesian_filter}
\begin{aligned}
    P(\hat{\theta}_{n},{\hat\sigma}_{\theta} ) &= \eta \cdot P(\hat{\theta}_{n},{\hat\sigma}_{\theta} ) ~\mathcal{N}(\theta_n, \sigma_{\theta}),\\
    \sigma_{\theta} &= \sigma_{0} + \lambda \cdot d,
\end{aligned}
\end{equation}
% \begin{equation}\label{eq_bayesian_filter}
% \begin{aligned}
%     \mathcal{N}(\theta_{n},\sigma_\theta ) &= \eta \cdot p(\hat{\theta}_{n},{\sigma}_{\hat\theta} ) p(\theta_n, \sigma_{\theta}),\\
%     \sigma_{\theta} &= \sigma_{0} + \lambda \cdot d,
% \end{aligned}
% \end{equation}
\kike{
    where $\mathcal{N}(\theta_n, \sigma_{\theta})$ is a normal distribution defined by $\theta_n$ as the angular orientation directly computed from $\textbf{n}$ and $\sigma_\theta$ as the variance associated with $\theta_n$ (i.e., uncertainty in the plane orientation); $d$ is the plane distance; $\lambda$ is a constant term that adjust $\sigma_\theta$ as function of $d$; $\sigma_0$ is the constant inherent noise in the measurement; lastly, $\eta$ is a term that normalizes the values into a probability function.
}

\kike{
    % In our implementation, to update $p(\hat{\theta}_{n},{\sigma}_{\hat\theta})$ with a new measurement~$\theta_{n}$, we assert that~$|\hat{\theta}_{n}-\theta_n|\leq \pi/3$, assuming that $p(\hat{\theta}_{n},{\sigma}_{\hat\theta})$ has been previously initialized.
    \ycheng{
    % In our implementation,
    \TODO{In practice, }given a new measurement $\theta_{n}$, we update the associated $P(\hat{\theta}_{n},{\hat\sigma}_{\theta})$ that satisfies $|\hat{\theta}_{n}-\theta_n|\leq \pi/3$.}
    \justin{
    If no posterior distribution satisfies the condition, we initialize $P(\hat{\theta}_{n},{\hat\sigma}_{\theta})$ with $\mathcal{N}(\theta_n, 2\pi)$. In the end, only estimations with $\hat\sigma_{\theta} < \pi/10$ are considered as plane orientations for the room.}
    }
\begin{figure}[t]
\centering
\includegraphics[width=1\linewidth]{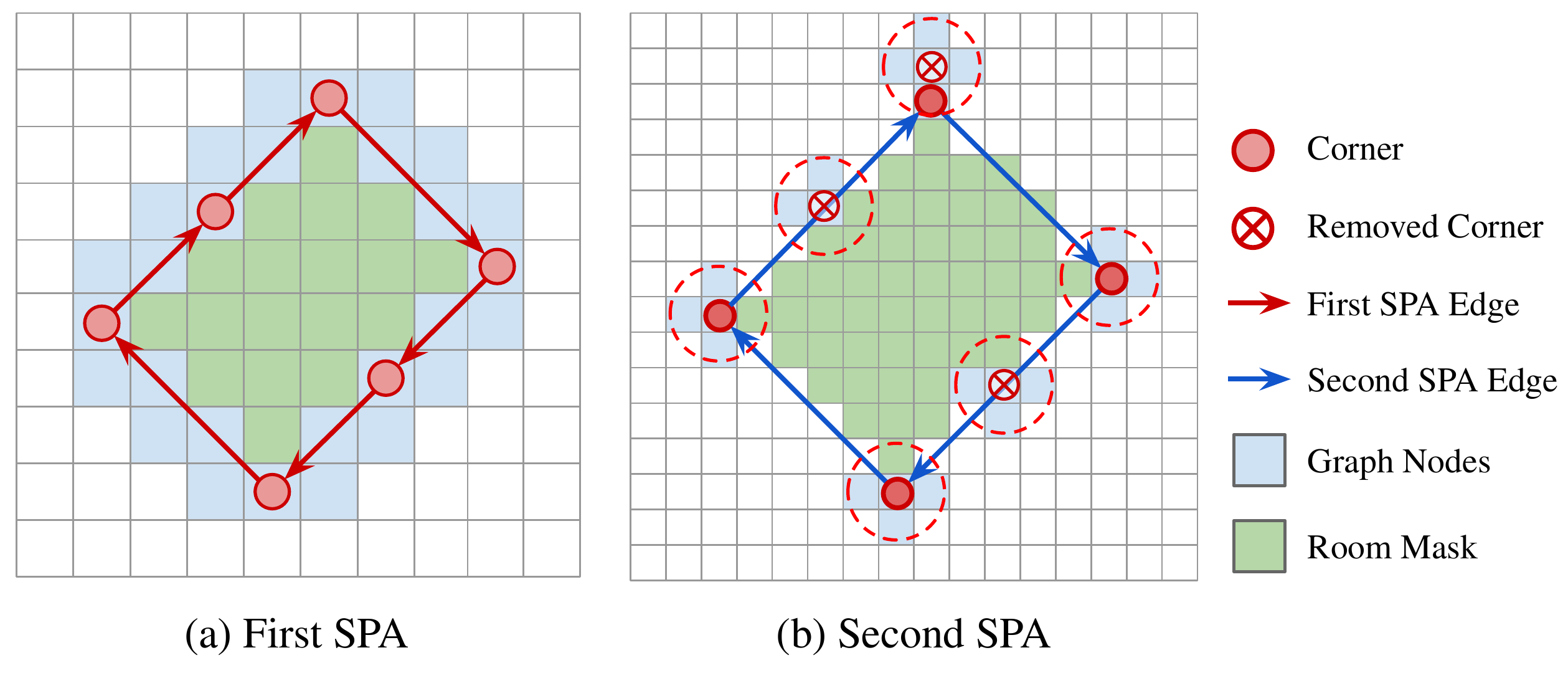}
% \includesvg[width=1\linewidth]{FIG/teaser.svg}
\vspace{-7mm}
\caption{\textbf{Iterative Shortest Path Algorithm (iSPA)}. We show two SPA evaluations as an example. (a) The first SPA estimates room corners around the room mask (green) under a coarse grid constrained by a maximum edge length, which will significantly speed up the optimization but also introduce redundant corners. (b) The second SPA evaluation searches over the neighboring areas around previous corner predictions (blue) with a finer grid size, which will help reduce redundant corners and refine the corner position. See Section~\ref{sec:Room-Shape Opt} for details.
% \TODO{Refer back to section.}
}
\label{fig:ispa}
\vspace{-4mm}
\end{figure}
\begin{figure*}[t!]

% \TODO{TO BE CHANGED.\\}
% \quad\quad\quad\quad\quad Ground Truth
% \quad\quad\quad\quad\quad\quad 360-DFPE
% \quad\quad\quad\quad\quad\quad\quad Floor-SP~\cite{chen2019floorSP}
% \quad\quad\quad\quad\quad\quad FloorNet~\cite{liu2018floornet}

\centering
\includegraphics[width=0.80\linewidth]{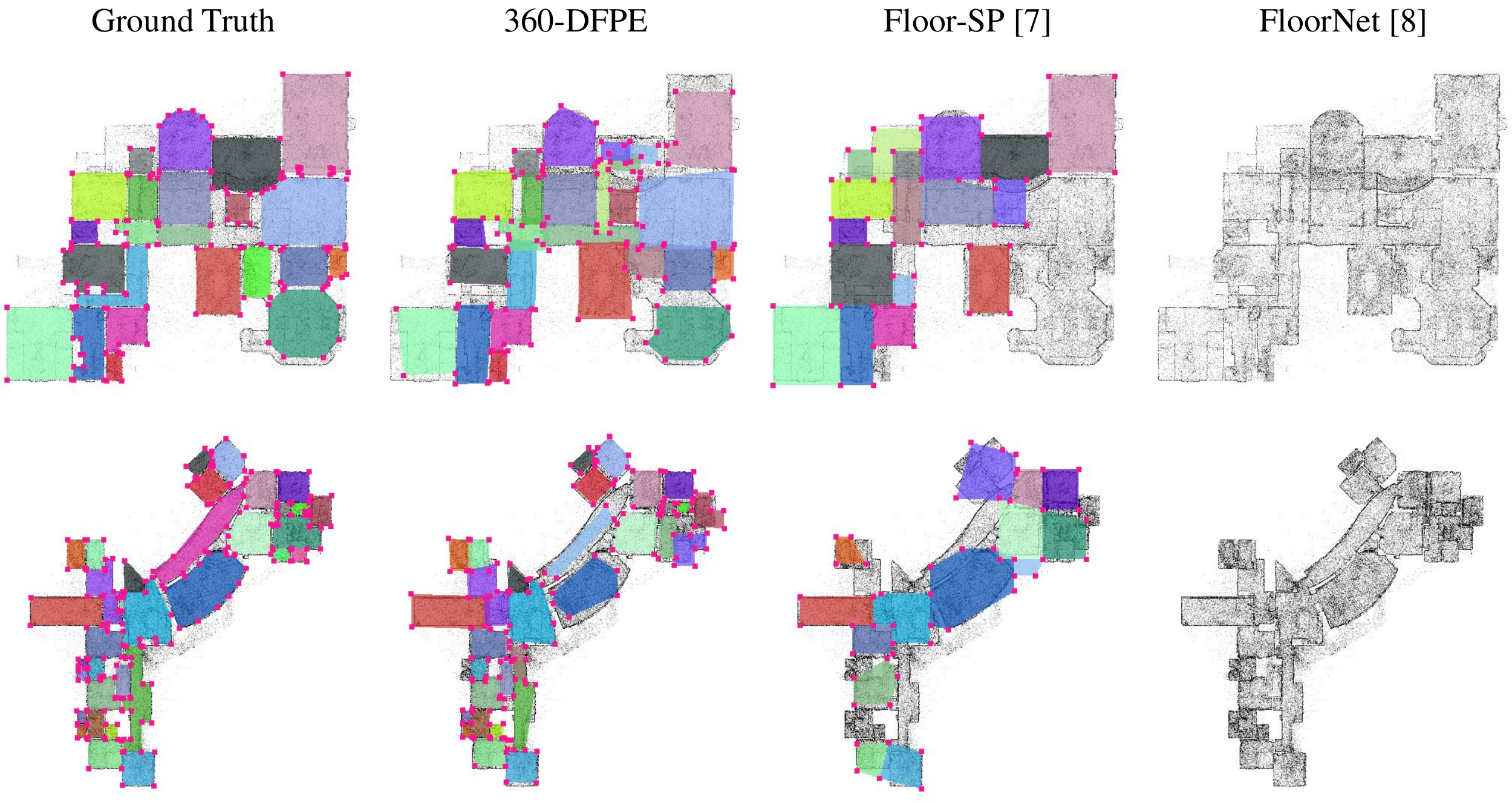}

% \includegraphics[width=0.9\linewidth]{FIG/results.png}
% \fbox{\rule{0pt}{6in} \rule{.9\linewidth}{0pt}}

% \includegraphics[width=0.9\linewidth,height=15cm]{Placeholder.png}
\vspace{-2mm}
\caption{
\textbf{Qualitative Comparison on MP3D-FPE.}
\justin{
% Draft
    % , and a point cloud-projected background to facilitate understanding.
    Compared with FloorNet~\cite{liu2018floornet} and Floor-SP~\cite{chen2019floorSP}, our 360-DFPE performs significantly better in two large scenes, which consist of a lot of rooms. 360-DFPE is able to successfully identify and reconstruct complex room shapes.
    % We find that achieves even better results in capturing the overall structure without relying on active sensors; moreover, the proposed method is more robust to diverse room shapes. In the two selected scenes above, FloorNet failed to reconstruct the whole scene, while our method is able to recover more and higher quality rooms than Floor-SP. 
    We use colors to represent room instances and magenta dots to highlight room corners.
    % However, we also notice that our method might wrongly identifies rooms if the boundaries between rooms are ambiguous, as shown in the bottom right rooms in first row.
% Note that the underneath black dots are point clouds to facilitate understanding, but they are used only in FloorNet and Floor-SP. 
}
% \ycheng{
    % \textbf{Room Identification and Room-shape Optimization.}
    % \textbf{(a-b) Room Identification.} To cluster the camera poses and the corresponding layouts that are within the same room, we project the piece-wise planes (orange) and camera poses (pink) onto a 2D grid.
    % If the edge between a new camera pose $C_{n+1}$ and the room center $O_R$ is cross any plane, we consider that the room is completed and group the camera poses within the room.
    % \textbf{(c-d) Room-shape optimization.} Inspired by \cite{chen2019floorSP}, we perform optimization over the room shape represented by corners $\mathbf{p}_i$ based on the information of planes and filtered wall orientations $\hat{\theta}$ of the room by Shortest Path Algorithm (SPA). Note that some of the rooms are not entered to be identified.
% }
}\label{fig:exp_fig}
\vspace{-4mm}
\end{figure*}
\vspace{-5mm}
\subsection{Room Shape Optimization}\label{sec:Room-Shape Opt}

\ycheng{
    % We enhance the shortest path algorithm (SPA) in prior works~\cite{chen2019floorSP, cabral2014piecewise} by introducing a coarse-to-fine and iterative strategy, reducing the run-time and increasing the accuracy.
    At the final stage, we introduce the room shape optimization to convert previous estimations into room shapes (i.e., a set of room corners). To optimize over the number of corners and their locations, we introduce
    an iterative coarse-to-fine strategy in the shortest path algorithm (SPA) proposed in prior works~\cite{chen2019floorSP, cabral2014piecewise}, where our method reduces the run-time and increases the accuracy. We name it as the iterative shortest path algorithm (iSPA).
    % The main idea of the SPA is to formulate the room shape optimization as solving the shortest path problem for a graph.
    % The graph considers potential corner locations as nodes, and the edges are weighted by a cost function based on the geometric evidence (e.g., plane-wall features, plane orientations, room masks). Therefore, the resulting nodes are the position of the room corners.
    \kike{
    The main idea of the shortest path algorithm for room shape estimation is to represent the room structure as a graph of possible corners as nodes and edges weighted by a cost function based on geometry evidence (e.g., plane-wall features, plane orientations, room masks). Therefore, by optimizing the shortest path, the resulting nodes are the corner positions of the room.}
    % The resulting nodes are the polygon vertices of the vectorized room shape.
    % yielding the minimal total cost.
}
    
% We enhance the SPA formulation in prior works~\cite{chen2019floorSP, cabral2014piecewise} by adding a coarse-to-fine and iterative design, reducing the run-time and increasing the accuracy. The ide

% First, we collect the planes within the room and project them into 2D density maps as $\mathcal{M}_{pl}$ with size $H \times W$.
% We also convert the room density map $\mathcal{H}$ into the binary room mask $\mathcal{M}_m$ by thresholding with a constant. 
% Then, we construct a graph $\mathcal{G}=(\mathcal{V}, E)$, where $\mathcal{V}$ is composed of the pixels in the 2D grid. Since setting all the pixels as nodes is infeasible, we select only the pixels in the neighboring region from the room mask boundary as valid nodes for $\mathcal{V}$. Given an edge connected by two pixels $\mathbf{x}=(u_1, v_1)$, $\mathbf{y}=(u_2, v_2)$, we define three cost functions for the edge weight based on the edge orientation, planes, and the room mask respectively:
\kike{
    In our design, after the camera leaves a room, we collect all the point sets $\mathbf{S}_0, \dots, \mathbf{S}_k$ in the room that define valid plane-wall features (see Section~\ref{sec:Plane Estimation}). Then, we project them into a 2D density map $M_{P}$. 
    \justin{We also binarize the room density function $H$ into a room mask $M_H$ with a threshold $0.5$~(see Section~\ref{sec:Room ID}).}
    % We also convert the room density function $\mathcal{H}$ (See Section~\ref{sec:Room ID}) into a binary room mask $\mathcal{M}_\mathcal{H}(u, v)$ with a threshold of $0.5$. 
    Lastly, we construct a weighted graph $\mathcal{G}=(\mathcal{V}, \mathcal{E})$, where $\mathcal{V}$ are the nodes composed by the neighboring pixels around $M_H$, and $\mathcal{E}$ are the edges.}

% \ycheng{
%     Given the planes and patches of a room, we first project them into a 2D grid map as $\mathbf{M}_{pl}$, $\mathbf{M}_{pc}$, both with a fixed size $H \times W$. The idea is to consider every pixel as a candidate location for a room corner. Therefore, we construct a graph which takes pixels as nodes and connect them with weighted edges. The weights are designed to be cost functions constrained by the previous estimations (e.g. planes, orientations, patch). We then apply the ordinary shortest path algorithm to retrieve a path, composed by a list of corners, with the minimal cost globally that best represents the room shape on the 2D projection.
% }
% \ycheng{
    % Here, we define the graph $\mathcal{G}=(\mathcal{V}, E)$.
    % Since setting all the pixel as nodes is infeasible, we set only the pixels in the dilated region from the patch's boundary as valid nodes for $\mathcal{V}$. We connect the edges that follow the major orientations $\mathbf{M}_\theta$ with a small angular tolerance. Given an edge connected by two pixels $\mathbf{x}=(u_1, v_1)$, $\mathbf{y}=(u_2, v_2)$, we define three cost functions for the edge weight based on the edge orientation, planes, and patches:

\kike{
    For the weight of an edge from $\mathbf{p}=(u_1, v_1)$ to $\mathbf{q}=(u_2, v_2)$, we define three cost functions:}
\begin{gather}
    \mathcal{L}_\text{ori} (\mathbf{p}, \mathbf{q}) = \min_{{\theta} \in \mathbf{M}_\theta} \Big|
        \arccos  \big(
        \mathbf{n}_{{\theta}} \cdot 
    % \mathbf{u}_{\mathbf{xy}}
    % u_{\mathbf{xy}}
        \frac{\overrightarrow{\mathbf{pq}}}{\| \overline{\mathbf{pq}} \|}
        \big)
    \Big| , \\
    \mathcal{L}_\text{plane} (\mathbf{p}, \mathbf{q}) = 
    \sum_{\mathbf{z}\in {E}(\mathbf{p}, \mathbf{q})} (1 - M_{P} (\mathbf{z})), \\
    \mathcal{L}_\text{mask} (\mathbf{p}, \mathbf{q}) = 
   \sum_{\mathbf{z}\in {E}(\mathbf{p}, \mathbf{q})} {M}_{H}(\mathbf{z}),
\end{gather}
% \kike{
%     where $\mathbf{M}_\theta$ is the set of wall-plane orientations $\mathbf{n}_{\hat{\theta}}$ for the room, and $\mathcal{E}(\mathbf{x}, \mathbf{y})$ is the set containing pixels on the line segment $\overline{\mathbf{xy}}$. We assign the edge weight by combining the cost functions with weighting parameters:
% }
% where $\mathbf{n}_{\hat{\theta}}$ is the normal vector of the orientation, $\mathbf{M}_\theta$ is the set of plane orientations for the room, and $\mathcal{E}(\mathbf{x}, \mathbf{y})$ is the set containing pixels on the line segment $\overline{\mathbf{xy}}$. We assign the edge weight by combining the cost functions with weighting parameters:
where $\mathbf{M}_\theta$ is the set of likely plane orientations for the room (see Section~\ref{sec:Plane Estimation}), $\mathbf{n}_{{\theta}}$ is the normal vector of the orientation, and $E(\mathbf{p}, \mathbf{q})$ is a set containing pixels on the line segment $\overline{\mathbf{pq}}$. Finally, we define our edge weight by the following:
\begin{multline}\label{eq_loss_spa}
    \mathcal{L}(\mathbf{p}, \mathbf{q}) =
    \lambda_\text{ori} \mathcal{L}_\text{ori} (\mathbf{p}, \mathbf{q})
    + \lambda_\text{plane}  \mathcal{L}_\text{plane}(\mathbf{p}, \mathbf{q}) \\
    + \lambda_\text{mask} \mathcal{L}_\text{mask}(\mathbf{p}, \mathbf{q})
    + \lambda_\text{complex},
\end{multline}
where $\lambda_\text{*}$ are weighting constants, and $\lambda_\text{complex}$ controls the model complexity (i.e., number of the room corners).
% For the starting edge, we sample two end-points of $\mathbf{S}_k$ set as the start-node and end-node. 
To prevent SPA from finding trivial solutions, we adopt the containment constraint proposed in \cite{chen2019floorSP}. \kike{By solving the shortest path problem, we can estimate the room corners; hereinafter, we refer to this estimation as a SPA evaluation.}
% \ycheng{
% % Some discussion about SPA
%     Note that the grid resolution is the trade-off between speed and precision. With higher resolution, the number of nodes and edges will be too large, and thus the computation is time-consuming. On the other hand, with lower resolution, the quantization error would affect the precision of the room corners.
% }
% \ycheng{
%     Compared to \cite{chen2019floorSP}, we simplifies formulation without considering the neighboring rooms. In addition, we enforce the plane orientation estimated from the data instead of following the Manhattan-world assumption, which allows us to handle diverse room shapes.
% }

\kike{
    The grid resolution of the density maps is correlated with the number of nodes and edges in the graph, which controls the computation time. Therefore, we propose a first SPA evaluation that uses a coarse grid and constrains edges with a maximum length. This limited length strategy significantly reduces the time complexity of building graph: from $O(N^2)$ to $O(\alpha N)$ where $\alpha \ll N$. However, it also introduces redundant estimated corners (see Fig.~\ref{fig:ispa}-(a)).
    }
    
\kike{
    To reduce the redundant corner estimations, as shown in Fig.~\ref{fig:ispa}-(b), we apply additional SPA evaluations using finer resolutions without the edge-length constraint, in where the nodes in $\mathcal{V}$ are now defined by neighboring pixels around the previous estimated corners.
    % In our implementation, we use regions around each previous estimated corner within 4 pixels.
    \TODO{
    In our implementation, we use 3 rounds of SPA with grid sizes \{64, 96, 96\} in a coarse-to-fine fashion for room shape optimization.
    }
}
\ycheng{
    In practice, iSPA not only removes redundant corner estimation from previous results but also reduces quantization error (i.e., the error due to discretization over a coarse grid), and increases the corner precision while having a faster run-time than the original SPA~\cite{chen2019floorSP}.
    % \TODO{ Note that we can adjust grid sizes or the number of SPA evaluations as a trade-off between efficiency and the quality of the room shape. 
    To further elaborate, we provide an ablation study of iSPA in Section~\ref{subsec:exp_ispa}.
    %The first SPA with restriction costs less time than the original one but it will enlarge the corner error at the same time. In practice, we can
}
\vspace{-1mm}
% \TODO{Let's summarize all in subsection, describing the whole pipeline and liking some details as: Room references, room creation, VO-Scale as visual odometry scale...etc etc}

\section{Experiments}
% Structure
% 1. Dataset
% 2. Metric
% 3. Main result
%   a. How to get gt_scale
% 4. Ablation study
%   a. Layout
%   b. HohoNet
%   c. Room ID (optional)
\ycheng{
    In this section, we first introduce our experimental setups, including datasets, evaluation metrics, and implementation details. Then, we provide our experimental results comparing with baselines and the ablation study.
}
% \vspace{-1.5mm}
%
% \justin{
%     In this section, we present our experiments on floor plan estimation. First, we will introduce our new collected dataset which contains 360 image sequences and corresponding point clouds. Secondly, we compare our method with the two state-of-the-art methods, Floor-Net~\cite{liu2018floornet} and Floor-SP~\cite{chen2019floorSP}. Moreover, we conduct several ablation studies to further investigate the effectiveness of our proposed components.
% }
%
%
% \TODO{In this section, qualitative and quantitative results are presented to validate the our proposed solution. For comparison purposes, the selected baselines are the current-state-of-art methods: Floor-SP\cite{chen2019floorSP} and Floor-Net~\cite{liu2018floornet}. Note that both methods rely on active sensors to build their input point cloud data of the entire scene, while our proposal assumes a sequence of 360-images as input. For fair comparisons we collect our own dataset where depth ground truth information is collected together with sequence of images. Thus this can be used to produce a clean point cloud of the entire scene following the description detailed in ~\cite{chen2019floorSP, liu2018floornet}. Both baselines are evaluated using these point cloud inputs using the original setting released in their implementations.}
\vspace{-1mm}
\TODO{\subsection{Dataset}\label{sec_dataset}}
    Since image sequences are not available in previous floor plan datasets \cite{liu2018floornet,chen2019floorSP,lin2019floorplanPartialRGBD,xiao2014reconstructing, cabral2014piecewise}, we collect and annotate a new dataset, MP3D-FPE. We synthesize users scanning through the scenes in Matterport3D~\cite{chang2017matterport3d} with the Minos simulator \cite{savva2017minos}.
    \justin{
    % Previous floor plan datasets lack of either image sequence \cite{liu2018floornet,chen2019floorSP}, or 360 images \cite{lin2019floorplanPartialRGBD,xiao2014reconstructing, cabral2014piecewise}. Therefore, we collect and annotate a new dataset, MP3D-FPE, by synthesizing users scanning through the scenes in Matterport3D~\cite{chang2017matterport3d} with Minos~\cite{savva2017minos}, the simulation environment. 
    MP3D-FPE contains 50 scenes and 687 rooms in total. For each scene, we provide the point cloud, sequence of panorama images with depth, camera pose information, and the floor plan annotation. This enables us to evaluate methods consuming various input modalities. As shown in Table~\ref{tab:dataset}, MP3D-FPE has larger scenes and more rooms per scene, making it more challenging compared to the previous floor plan dataset proposed by Floor-SP~\cite{chen2019floorSP}.
    % , e.g., non-Manhattan room-shapes, large-scale scenes, etc.
    % See  for more information.
}

\begin{table}[thbp]
\fontsize{10}{16}\selectfont
\vspace{-2mm}
\centering
\caption{Statistics of MP3D-FPE Dataset.}
\resizebox{\columnwidth}{!}{
\begin{tabular}{c c c cc c cc}
\hline

\multirow{2}{*}{Dataset} &
\multirow{2}{*}{No. Scenes} &
\multicolumn{1}{c}{} &
\multicolumn{2}{c}{No. Rooms} &
\multicolumn{1}{c}{} &
\multicolumn{2}{c}{Area ($m^2$)} \\
% \hline
\cline{4-5}
\cline{7-8}
\multicolumn{1}{c}{} &
\multicolumn{1}{c}{} &
\multicolumn{1}{c}{} &
\multicolumn{1}{c}{avg} &
\multicolumn{1}{c}{std} &
\multicolumn{1}{c}{} &
\multicolumn{1}{c}{avg} &
\multicolumn{1}{c}{std} \\ 

\hline
Floor-SP~\cite{chen2019floorSP} & \textbf{99} & & 7.07 & 2.26 & & 104.52 & 44.36 \\
MP3D-FPE (Ours) & 50 & & \textbf{13.74} & \textbf{9.00} & & \textbf{685.72} & \textbf{760.52} \\
\hline
\end{tabular}
}
\label{tab:dataset}
\vspace{-5mm}
\end{table}

\begin{table*}[htbp]
\fontsize{7}{10}\selectfont
\vspace{2mm}
\centering
\caption{Evaluation Result on MP3D-FPE. SR-*\% represents different ratio of keyframes used in relative scale recovery.}
\resizebox{\textwidth}{!}{
\begin{tabular}{ l c c c c cc c cc c cc c cc}

\hline

% First header
\multirow{2}{*}{Method} &
\multirow{2}{*}{Input} &
\multicolumn{1}{c}{} &
\multicolumn{2}{c}{Room IoU$_{0.3}$} &
\multicolumn{1}{c}{} &
\multicolumn{2}{c}{Room IoU$_{0.5}$} &
\multicolumn{1}{c}{} &
\multicolumn{2}{c}{Room IoU$_{0.7}$} &
\multicolumn{1}{c}{} &
\multicolumn{2}{c}{Corner} \\
\cline{4-5}
\cline{7-8}
\cline{10-11}
\cline{13-14}

\multicolumn{2}{c}{} &
\multicolumn{1}{c}{} &
\multicolumn{1}{c}{Recall} &
\multicolumn{1}{c}{Precision} &
\multicolumn{1}{c}{} &
\multicolumn{1}{c}{Recall} &
\multicolumn{1}{c}{Precision} &
\multicolumn{1}{c}{} &
\multicolumn{1}{c}{Recall} &
\multicolumn{1}{c}{Precision} &
\multicolumn{1}{c}{} &
\multicolumn{1}{c}{Recall} &
\multicolumn{1}{c}{Precision} \\
\hline

FloorNet~\cite{liu2018floornet} & Point Cloud & &
19.82 & 48.61 & &
11.60 & 20.84 & &
6.53  & 12.37 & &
15.23 & 66.69 \\

Floor-SP~\cite{chen2019floorSP} & Point Cloud & &
72.07 & 76.16 & &
61.39 & 61.87 & &
46.20 & 43.34 & &
% 38.90 & \textbf{76.70} \\
53.58 & 76.27 \\
\hline
360-DFPE (SR-20\%) & 360-Images & &
% 79.58 & 79.19 & & 
% 68.68 & 68.53 & &
% 46.07 & 46.76 & &
% 51.39 & 64.60 \\
79.82 & 83.50 & & 
68.33 & 71.09 & &
45.25 & 47.72 & &
\textbf{67.79} & 69.82 \\
360-DFPE (SR-100\%) & 360-Images & &
% \textbf{85.60} & \textbf{87.90} & &
% \textbf{76.33} & \textbf{77.85} & &
% \textbf{55.28} & \textbf{56.73} & &
% \textbf{69.71} & \textbf{78.33} \\
\textbf{83.02} & \textbf{90.99} & &
\textbf{72.01} & \textbf{78.21} & &
\textbf{52.09} & \textbf{56.40} & &
67.66 & \textbf{78.90} \\
% \textbf{55.53} & 74.26 \\
\hline

\end{tabular}
}
\label{tab:exp_main}
\vspace{-4mm}
\end{table*}

% \justin{
%     Our collected dataset contains 50 scenes from Matterport3D~\cite{chang2017matterport3d} and rendered by Minos~\cite{savva2017minos} to generate sequential panorama frames, in both RGB and depth. In contrast to the dataset proposed along with Floor-SP~\cite{chen2019floorSP}, our dataset has a wider diversity and more challenging scenes, e.g., non-Manhattan room-shapes, large-scale scenes, etc. Comparison with dataset from Floor-SP is shown in Fig. XXX. Corner and room labels are provided for the evaluation purpose.
% }

\subsection{Metrics}\label{sec_metrics}
% 1. corners (a) PR 256*256 grid (2) L2 error
% 2. Room IoU
% 3. scale recovery
\justin{
    Following FloorNet~\cite{liu2018floornet} and Floor-SP~\cite{chen2019floorSP}, we evaluate the floor plan results with corners and room metrics, which are defined as the follows:
}
\subsubsection{Corner}
\ycheng{
   For the corner metric, we evaluate the recall and precision of the estimated corners. First, we project all corners in the scene onto a 2D grid with size $256 \times 256$. Then, each predicted corner is determined successfully reconstructed if there is a ground-truth corner within 10 pixels. When multiple predictions are around one ground truth, only the closest one is counted as true-positive; the rest are treated as false-positive cases.
    % A ground-truth corner is determined successfully reconstructed if there is a corner prediction within 10 pixels. For multiple corner predictions, only the closest corner to the ground-truth is counted as true-positive, the rest is treated as false-positive cases. 
    % In addition, only the closest corner will be counted as true-positive when multiple predictions are around one ground-truth corner, while the rest will be treated as false-positive cases.
}
% \TODO{
%     We report the precision and recall. Also, because projecting the corners on a fix 2D grid will lose the real-world scale, especially for the large scenes, we also propose to use L2 error on original scale as a metric to evaluate the predicted corners. The L2 error is computed by the Chamfer distance between ground truth and estimated room corners. 
% }
\subsubsection{Room}
\ycheng{
    Like the corner metric, we report the recall and precision values in terms of the room area. An estimated room is considered successfully reconstructed if the intersection-over-union (IoU) with the ground-truth is above a certain IoU threshold.
    % In this paper, we consider three different IoU thresholds, 0.3, 0.5 and 0.7.
    % In contrast to the corner metric which consider the low-level precision of the floor plan, the room metric measures the quality of the reconstructed rooms, which are the basic units of floor plan.
}
\vspace{-1mm}
\subsection{Experimental Results}\label{sec_experiments_results}
\kike{
    % To corroborate the performance of the proposed 360-DFPE, we compare it with two state-of-the-art baselines\footnote{We do not compare with MonteFloor~\cite{stekovic2021montefloor} since the official code is not available during the submission of this manuscript.} Floor-SP~\cite{chen2019floorSP} and Floor-Net~\cite{liu2018floornet}, using MP3D-FPE. 
    We compare 360-DFPE with two state-of-the-art baselines\footnote{We do not compare with MonteFloor~\cite{stekovic2021montefloor} since the official code is not available during the submission of this manuscript.}, Floor-SP~\cite{chen2019floorSP} and Floor-Net~\cite{liu2018floornet}, on our MP3D-FPE dataset. 
    For reference purposes, we present two different implementations, using only initial 20\% of keyframes and using all keyframes for scale recovery (see Section~\ref{sec:Scale Recovering}).
    % : (20\%) and (full), where the former uses only 20\% of the data for recovering the missing scale $S_{VO}$, and the latter uses all the data (cf. Section~\ref{sec:Scale Recovering}). 
    % In addition, we implement 3 iSPA iterations with resolutions \{64, 96, 96\} (see Section~\ref{sec:Room-Shape Opt}).
    \ycheng{
    Moreover, since the ground-truth floor plans are labeled on top of the point cloud, we \TODO{align} our predicted floor plan to the real scale of the point cloud for a fair comparison. Therefore, during the evaluation, we align the estimated poses with ground truth poses to recover the real scale.
    }}

% \kike{
%     The qualitative result of floor plan estimation is presented in Table~\ref{tab:exp_main}, wher our 360-DFPE significantly outperforms the two baselines in both room and corner metrics.
%     Note that even using only $20\%$ of the data for scale recovering is enough to generally evaluate a better estimation than the proposed baselines.
% }
% 
% \kike{
%     For room evaluation, under different IoU thresholds, our method achieves the best recall and precision results. By considering room recall, we can assert that our our proposed room identification~(see Section~\ref{sec:Room ID}) estimates room instances in the floor plan with higher success rate than the baselines. Based on the room precision results, we can assert that our room mask definition (which is also based on our room identification) together with the proposed iSPA are capable of recovering a better room shape geometry compared with the baselines.
% }
The quantitative result of floor plan estimation is presented in Table~\ref{tab:exp_main}, where our 360-DFPE significantly outperforms the two baselines. For the room metric, our method achieves the best recall and precision results under different IoU thresholds, which shows that our proposed pipeline with novel room identification and iSPA can estimate the room instances with higher success rates and more accurate room shapes. Note that our better room geometry estimation also leads to significantly higher corner recall.

On the other hand, comparing the number of keyframes used in scale recovery, using all the data performs slightly better than using only the initial 20\% in both corner precision and room metrics. This is because using all data 
% for relative scale recovery 
considers dynamic factors (e.g., different room sizes, changes in the camera height, and small drift in odometry estimation) and hence estimates a more robust scale. Nevertheless, using only the initial 20\% keyframes can already acquire a good scale $s$, which achieves better results than the baselines without requiring the whole scene data in advance.
% \kike{
%     Based on the corner results, we corroborate that our proposed formulation is capable of estimating room corners with lower false-positive and more accurate estimations, i.e., a better room topology estimation. Note that using all the data for scale recovering slightly outperforms our version that uses only $20\%$. This is because, by using all data, our scale recovering formulation can take into account all the scene conditions (e.g., different room sizes, changes in the camera height, and small drift in odometry estimation) to align coherently the whole scene in advance. However, by using only the first $20\%$, we can still acquire a good enough scale $S_{VO}$ showing competitive results than the proposed baselines, without requiring the whole data in advance.
% }

% \kike{
%     The quantitative evaluation is presented in Table.~\ref{tab:exp_main}.   
% }
% 
% Figure results shows:
% 1. No Manhatten world assumption -> good & axis-aligned constraint
% 2. Room ID -> good
% 3. Failure case?
\ycheng{
    We show visualization examples in Fig.~\ref{fig:exp_fig}. These scenes are large, consist of numerous rooms, and contain rooms with complex shapes. As a result, FloorNet~\cite{liu2018floornet} suffers from producing valid rooms in the floor plan due to its strong assumptions about the room shape.
    % ROOM ID
    On the other hand, Floor-SP~\cite{chen2019floorSP} predicts room instances on 2D density map with fixed size based on ground-truth point clouds. In some cases, it may be hard to detect room boundaries on the density map. Therefore, in both the first and second rows, Floor-SP fails to reconstruct several small rooms. On the contrary, our proposed room identification successfully differentiates the rooms by reasoning the room layout geometries and the camera poses within the scene.
    % Manhattan world
    Moreover, our method does not enforce prior assumptions such as Manhattan-world or require axis-aligned point clouds as input. Hence, 360-DFPE has the ability to handle rooms with more complex shapes compared to the baselines (see the second row of Fig.~\ref{fig:exp_fig}).
}

\vspace{-1mm}
\subsection{Ablation Study}\label{subsec:exp_ispa}

% \subsubsection{Layout Backbone}
% \ycheng{
%     To show the generalization ability to adopt different 360-layout algorithms, we replace the  HorizonNet~\cite{sun2019horizonnet} backbone with HohoNet~\cite{sun2021hohonet}. As shown in Tab.~\ref{tab:exp_layout}, the performance of using HorizonNet and HohoNet is similar, which means that our method does not have reliance on a certain 360-layout model.
% }
% \input{TAB_SRC/exp_layout}
\ycheng{
    To verify the effectiveness of iSPA, we provide a study in Table~\ref{tab:exp_spa}. As shown in the first row, the original SPA, which is similar to the way in Floor-SP~\cite{chen2019floorSP}, suffers from a long run-time. Enforcing fixed grid size can reduce the run-time (see the second row). In the \TODO{third} row, although constraining maximum edge-length further speeds it up, it will introduce redundant corners and lower the corner precision. In the fourth and fifth row, additional SPA evaluations in a coarse-to-fine fashion increase corner precision.
}

\begin{table}[htbp]
\fontsize{18}{26}\selectfont
\vspace{-2mm}
\centering
\caption{
\textbf{Ablation Study of ISPA.} We select the fourth row (3x iterations) for final version as a balance between corner precision and run-time.
}
\resizebox{\columnwidth}{!}{
\begin{tabular}{c c c c cc c cc c c}

\hline

% First header
\multicolumn{1}{c}{Fixed} &
\multicolumn{1}{c}{Limited} &
\multirow{2}{*}{Iteration} &
\multicolumn{1}{c}{} &
\multicolumn{2}{c}{Room IoU$_{0.5}$} &
\multicolumn{1}{c}{} &
\multicolumn{2}{c}{Corner} &
\multicolumn{1}{c}{} &
% \multirow{2}{*}{Time (s/r)} \\
\multicolumn{1}{c}{Time} \\
\cline{5-6}
\cline{8-9}

\multicolumn{1}{c}{Size} &
\multicolumn{1}{c}{Edge Length} &
\multicolumn{1}{c}{} &
\multicolumn{1}{c}{} &
\multicolumn{1}{c}{Recall} &
\multicolumn{1}{c}{Precision} &
\multicolumn{1}{c}{} &
\multicolumn{1}{c}{Recall} &
\multicolumn{1}{c}{Precision} &
\multicolumn{1}{c}{} &
\multicolumn{1}{c}{(Sec. / Room)} \\
\hline

% v9.1-iou5-merge_corners_0.5
 & & 1x & &
70.51 & 76.66 & &
70.48 & 71.03 & &
80.71 \\

% v9.1-iou5-s64-merge_corners_0.5
v & & 1x & &
72.28 & 78.84 & &
69.44 & 74.20 & &
29.83 \\

% v8.1-iou5-new_angle_w2-l15-s64-merge_corners_0.5
v & v & 1x & &
\textbf{72.39} & \textbf{78.91} & &
\textbf{74.83} & 54.01 & &
\textbf{9.92} \\

% v8.1-iou5-new_angle_w2-l15-s64-96-96-merge_corners_0.5
\hline\hline
v & v & 3x & &
72.01 & 78.21 & &
67.66 & 78.90 & &
34.28 \\
\hline\hline
% v8.1-iou5-new_angle_w2-l15-s64-96-96-128-128-merge_corners_0.5
v & v & 5x & &
71.83 & 78.22 & &
67.58 & \textbf{80.10} & &
49.03 \\

\hline

\end{tabular}}
\label{tab:exp_spa}
\vspace{-4mm}
\end{table}
\section{Failure Cases and Limitations}

\begin{figure}

\centering

\includegraphics[width=0.8\linewidth]{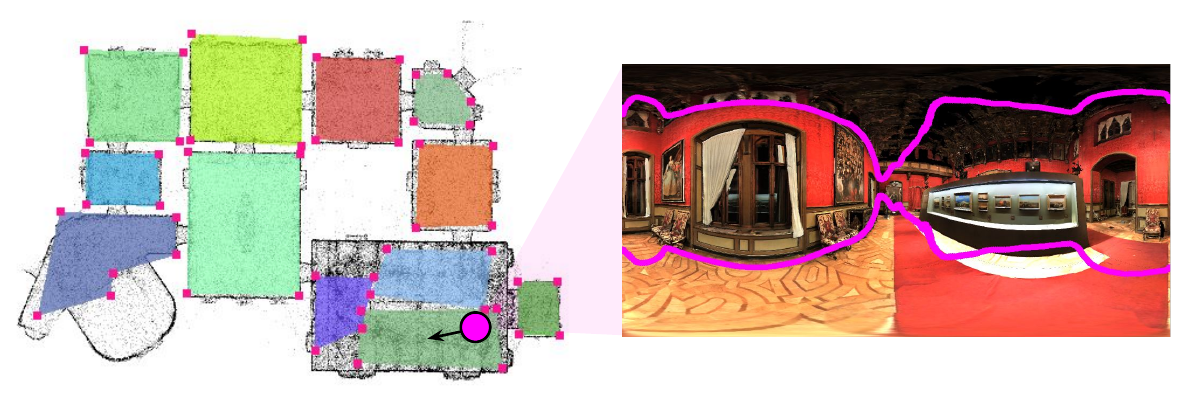}
\vspace{-3mm}
\caption{
\textbf{Failure Case.}
\ycheng{
% In the red box, we show the failure case that a large room is split into small parts due to noise from the layout backbone and the keyframes moving too fast.
We show that the noise from the 360-layout estimation could cause the room identification to fail, making a large room split into small parts. \TODO{Note that the floor is occluded by a large furniture.} 
}
}\label{fig:exp_fail}
\vspace{-4mm}
\end{figure}

\TODO{One of the common failure cases of 360-DFPE is due to error and drift produced by a monocular VSLAM, which may lead to a wrong layout registration in extreme cases, e.g., drastic movements and textureless regions.
% Especially when encountering textureless regions, the error of VSLAM may increase, 
% making it more difficult to recover reliable visual odometry scale. %hence constraining the whole system.
% In addition, noisy 360-layout estimations would affect the final floor plan estimation. This noise may be due to the occlusion of large furniture (e.g., Fig.~\ref{fig:exp_fail}) or atypical scenes, such as large halls or semi-open spaces.
In addition, noisy 360-layout estimations, due to the occlusion of large furniture (e.g., Fig.~\ref{fig:exp_fail}) or atypical scenes (e.g., large halls or semi-open spaces), may affect our final estimation. Lastly, since the proposed pipeline relies on a deep-learning method for layout estimation, it requires enough computation resources to run the pre-trained model. 
% On the other hand, 360-layout estimator could fail when the scene is noisy (e.g., occluded by large furniture) or for scenes unlike ordinary indoor rooms (e.g. large halls or semi-open spaces.
% For example, Fig.~\ref{fig:exp_fail} shows the failure caused by a large furniture in the middle of the room.
% Moreover, the proposed room identification procedure requires a smooth and continuous sequence of key-frames. Using sparse locations may lead to splitting false small rooms. 
% Moreover, since 360-DFPE uses keyframes as inputs, when the camera poses of those keyframes move too fast, a large room could be divided into several small parts because the layouts are disconnected.
% Another limitation of 360-DFPE is its dependency on the the number of keyframes registered per room. Thus, moving through the scene too fast may spawn few number of keyframes which may split large scene into multiple false small rooms.
% Lastly, although the modified iSPA is significantly faster than the previous approach, there is still room to improve towards real-time application.
}

\TODO{
    We believe that the mentioned limitations and failure cases may be addressed by using more robust solutions for both VSLAM and layout estimation.
    % We believe that the error caused by layout noise could be reduced by a more robust room layout backbone. 
    Moreover, closer integration between the VSLAM and floor plan estimation may increase the robustness of our current proposed pipeline.
    % We also believe that as the future direction, closer integration between the VSLAM system and floor plan estimation may mitigate the above-mentioned issues.
    % Currently, the drift issue is out of the scope of this paper. However, 
    % \TODO{TODO SPA is the bottleneck. Others can run in real-time, providing users feedback.}
}

% \section{Failure Cases and Limitations}
% One of the common failure cases of 360-DFPE is due to the error and drift produced by VSLAM. \TODO{This scenario might occur when encountering textureless regions. On the other hand, 360-layout estimator could fail when the scene is noisy (e.g., occluded by large furniture) or scenes unlike ordinary indoor rooms (e.g. large halls or semi-open spaces. For example, Fig.~\ref{fig:exp_fail} shows the failure caused by a large furniture in the middle of the room. In addition, since 360-DFPE uses keyframes as inputs, when the camera poses of those keyframes are moving too fast, a large room could be divided into several small parts because the layouts are disconnected.}
% \TODO{
%     Lastly, although the modified iSPA is faster than the previous approach, there is still a gap towards real-time application. 
% }

% \ycheng{
%     In the future, we believe that the error caused by layout noise could be reduced by a more robust room layout backbone. We also believe that as the future direction, closer integration between the VSLAM system and floor plan estimation may mitigate the above-mentioned issues.
%     % Currently, the drift issue is out of the scope of this paper. However, 
%     % \TODO{TODO SPA is the bottleneck. Others can run in real-time, providing users feedback.}
% }

\section{Conclusions}\label{sec:Conclusion}

In this work, we present 360-DFPE, a new direct floor plan estimation that uses a sequence of 360-images without relying on active sensors.
\kike{Our proposed solution can perform sequentially without strictly requiring the entire scene in advance. In particular, 360-DFPE consists of a novel 360-layout registration and a scale recovery, a novel room identification that sequentially tracks the corresponding room for each camera location leveraging only the surrounding geometry, and lastly, a novel coarse-to-fine SPA formulation that enhances previous solutions~\cite{cabral2014piecewise, chen2019floorSP} achieves higher speed and precision.}
% The pipeline of 360-DFPE can perform in a sequential manner without requiring the knowledge of the entire scene in advance (but only use initial 20\% keyframes for warm-up).
% Incorporating with visual SLAM system and 360-layout estimator, our novel 360-layout registration and scale recovery can align multiple layout estimations without additional assumptions.
% \TODO{In addition}, we propose the iterative shortest path algorithm which enhances previous method in terms of speed and precision.
% Using only images, Our 360-DFPE outperforms current state-of-the-art floor plan estimation methods in our newly collected dataset, MP3D-FPE.
Our 360-DFPE outperforms current state-of-the-art solutions by reconstructing floor plan structures regardless of the size, axis-alignment, and room topology of the scene.

% \TODO{For the benefit of the research community, our implementation and dataset (MP3D-FPE) will be released with this publication.}

% We present 360-DFPE, a direct floor plan estimation method that uses sequence of 360-images without relying on active sensors. Our approach leverages a loosely coupled integration between a monocular visual SLAM solution and a 360-room layout approach, estimating camera poses at each frame and projecting room layout geometry in the scene. In a sequential fashion, we register the 360-layout estimation for each keyframe, estimate rooms and planes, and finally optimize the room shape.
% \TODO{
% In our newly collected dataset with sequences of 360-images, our 360-DFPE out-performs current state-of-the-art floor plan estimation algorithms that rely on active sensors and acquire the entire scene data in advance. In the future....
% }

\section*{Acknowledgments}
This work is supported by the MOST Joint Research Center for AI Technology and All Vista Healthcare, Taiwan Computing Cloud, and MOST 110-2634-F-007-016. We thanks Professor Wei-Chen Chiu for his advises on this project.

% \vspace{-2mm}
% \section*{Acknowledgments}
% This work is supported by the MOST Joint Research Center for AI Technology and All Vista Healthcare, National Center for High-Performance Computing, and MOST 110-2634-F-007-016. We thanks Professor Wei-Chen Chiu for his advises on this project.

\bibliographystyle{IEEEtran}
\bibliography{reference}

% \newpage

% \section{Biography Section}
% If you have an EPS/PDF photo (graphicx package needed), extra braces are
%  needed around the contents of the optional argument to biography to prevent
%  the LaTeX parser from getting confused when it sees the complicated
%  $\backslash${\tt{includegraphics}} command within an optional argument. (You can create
%  your own custom macro containing the $\backslash${\tt{includegraphics}} command to make things
%  simpler here.)
 
% \vspace{11pt}

% \bf{If you include a photo:}\vspace{-33pt}
% \begin{IEEEbiography}[{\includegraphics[width=1in,height=1.25in,clip,keepaspectratio]{fig1}}]{Michael Shell}
% Use $\backslash${\tt{begin\{IEEEbiography\}}} and then for the 1st argument use $\backslash${\tt{includegraphics}} to declare and link the author photo.
% Use the author name as the 3rd argument followed by the biography text.
% \end{IEEEbiography}

% \vspace{11pt}

% \bf{If you will not include a photo:}\vspace{-33pt}
% \begin{IEEEbiographynophoto}{John Doe}
% Use $\backslash${\tt{begin\{IEEEbiographynophoto\}}} and the author name as the argument followed by the biography text.
% \end{IEEEbiographynophoto}

% \vfill

\end{document}